\documentclass{article}

\usepackage{microtype}
\usepackage{graphicx}
\usepackage{subfigure}
\usepackage{booktabs} 

\usepackage{hyperref}

\usepackage[algo2e]{algorithm2e} 

\usepackage[accepted]{icml2023}

\usepackage{amsmath}
\usepackage{amssymb}
\usepackage{mathtools}
\usepackage{amsthm}
\usepackage{multirow}
\usepackage{comment}
\usepackage{color, colortbl}
\usepackage{enumitem}
\usepackage{scalerel}

\usepackage[capitalize,noabbrev]{cleveref}
\usepackage[textsize=tiny]{todonotes}

\theoremstyle{plain}

\theoremstyle{definition}

\theoremstyle{remark}

\newcolumntype{P}[1]{>{\centering\arraybackslash}p{#1}}
\colorlet{shadecolor}{gray!30}
\definecolor{shamrockgreen}{rgb}{0.0, 0.62, 0.38}
\definecolor{urobilin}{rgb}{0.88, 0.68, 0.13}
\definecolor{sinopia}{rgb}{0.8, 0.25, 0.04}

\usepackage{xcolor,pifont}
\newcommand*\colourcheck[1]{%
  \expandafter\newcommand\csname #1check\endcsname{\textcolor{#1}{\ding{52}}}%
}
\colourcheck{shamrockgreen}
\colourcheck{urobilin}

\newcommand{\todoblue}[1]{\textcolor{blue}{#1}}
\definecolor{commentcolor}{RGB}{110,154,155}   
\definecolor{defcolor}{RGB}{225,81,145}

\newcommand{\PyComment}[1]{\footnotesize \ttfamily \textcolor{commentcolor}{\# #1}}  
\newcommand{\PyCode}[1]{\footnotesize \ttfamily \textcolor{black}{#1}} 
\newcommand{\PyDef}[1]{\footnotesize \ttfamily \textcolor{defcolor}{#1}} 

\icmltitlerunning{On the Importance of Feature Decorrelation in RL}

\begin{document}

\twocolumn[
\icmltitle{On the Importance of Feature Decorrelation for \\ 
            Unsupervised Representation Learning in Reinforcement Learning}




\begin{icmlauthorlist}
\icmlauthor{Hojoon Lee}{sch,comp1}
\icmlauthor{Koanho Lee}{sch}
\icmlauthor{Dongyoon Hwang}{sch}
\icmlauthor{Hyunho Lee}{comp1}
\icmlauthor{Byungkun Lee}{sch}
\icmlauthor{Jaegul Choo}{sch}
\end{icmlauthorlist}

\icmlaffiliation{sch}{KAIST}
\icmlaffiliation{comp1}{KaKao Enterprise}

\icmlcorrespondingauthor{Hojoon Lee}{joonleesky@kaist.ac.kr}

\icmlkeywords{Machine Learning, ICML}

\vskip 0.3in
]



\printAffiliationsAndNotice{} 

\begin{abstract}
Recently, unsupervised representation learning (URL) has improved the sample efficiency of Reinforcement Learning (RL) by pretraining a model from a large unlabeled dataset. 
The underlying principle of these methods is to learn temporally predictive representations by predicting future states in the latent space. 
However, an important challenge of this approach is the representational collapse, where the subspace of the latent representations collapses into a low-dimensional manifold.
To address this issue, we propose a novel URL framework that causally predicts future states while increasing the dimension of the latent manifold by decorrelating the features in the latent space.
Through extensive empirical studies, we demonstrate that our framework effectively learns predictive representations without collapse, which significantly improves the sample efficiency of state-of-the-art URL methods on the Atari 100k benchmark.
The code is available at \todoblue{\url{https://github.com/dojeon-ai/SimTPR}}.
\end{abstract}

\section{Introduction}
\label{introduction}

Deep Reinforcement Learning (RL) has made a significant advance in solving various  sequential decision-making problems \cite{dqn, alphago, gu2017deep, raghu2017sepsis, alphastar, nie2021treatment}.
The learning process of RL is inherently online, involving an iterative loop of data collection and policy optimization. 
However, 
in many real-world problems, the availability of online data collection is often limited as it can be expensive (e.g., robotics, and educational agents) or even dangerous (e.g., autonomous driving, and healthcare) \cite{offline_rl_survey, lecun_path}. To alleviate the burden of online data collection, researchers have explored the use of offline datasets by adopting the \textit{pretrain then finetune} paradigm in RL \cite{cpc, atc, sgi, lightweight_probe}.

In this paradigm, the agent trains an encoder from the offline dataset during the pretraining phase. Subsequently, during the finetuning phase, the agent leverages this encoder to efficiently optimize the policy for a downstream task with a limited number of online interactions.
When the action or reward labels are present in the offline dataset, pretraining is typically performed in a supervised manner, with the goal of predicting the labels for each state \cite{christiano2016idm, cql, bcq, decision_transformer}. 
Despite their simplicity, the acquisition of the labels is often limited, as it requires additional effort from human annotators.
As a result, recent studies have focused on developing unsupervised representation learning methods from the datasets that only consist of states, which are easily accessible, large, and do not require any additional labeling process ~\cite{rssm, atc, seo2022apv, baker2022vpt}.


\begin{figure}[t]
\begin{center}
\includegraphics[width=0.98\linewidth]{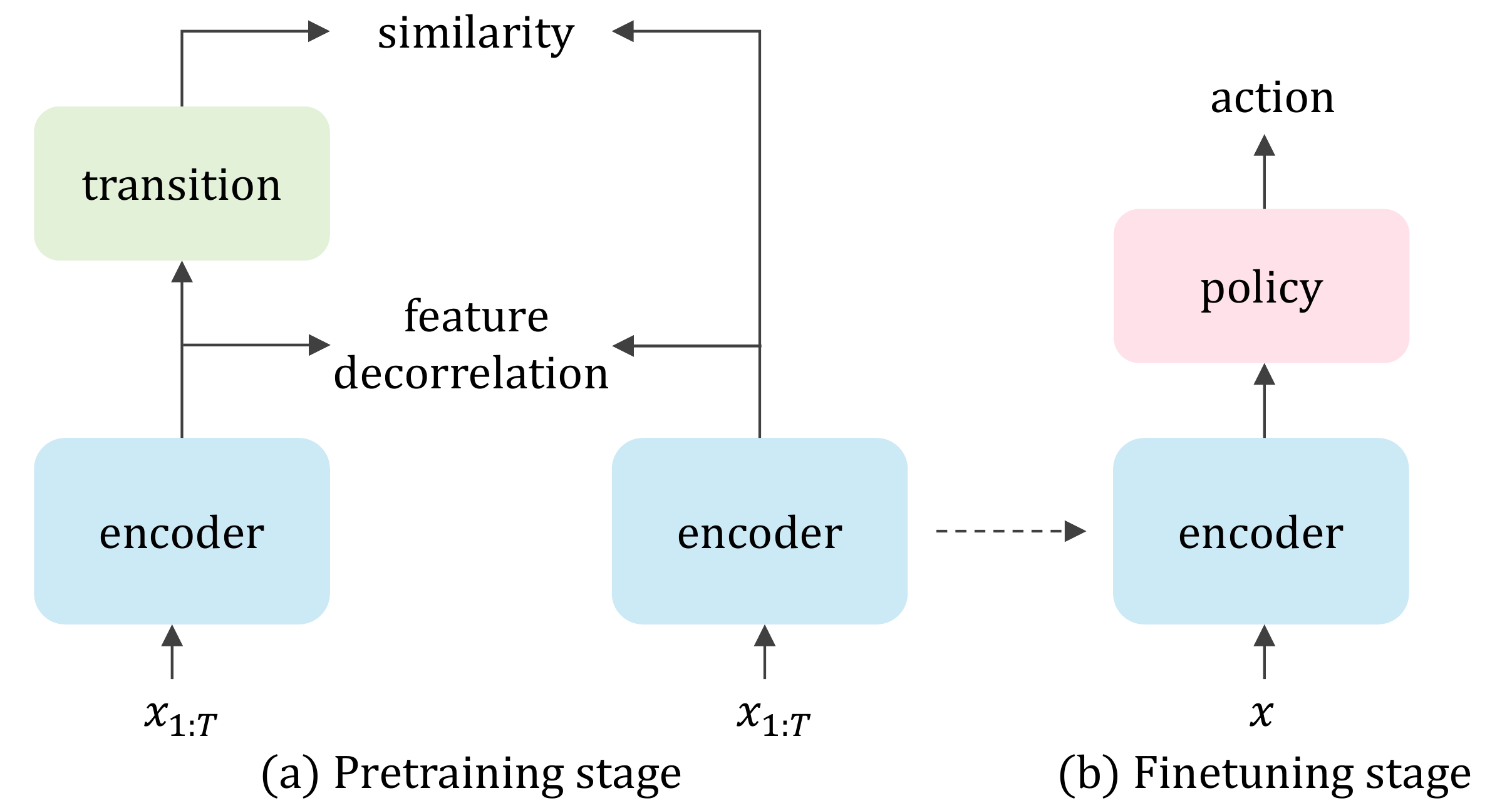}
\end{center}
\vspace{-3mm}
\caption{\textbf{SimTPR framework}. At pretraining, the encoder is trained to predict future states while decorrelating the encoded features in the latent space. Then, the pretrained encoder is used as a prior to optimize the policy at the finetuning stage.}
\vspace{-3mm}
\label{fig:intro}
\end{figure}

RL has a rich history of research on unsupervised representation learning from states, with seminal studies dating back to the 1990s and early 2000s ~\cite{dayan1993seminar, tesauro1995tdlearning, littman2001predictive, singh2003learning}.
A common principle across these studies is learning temporally predictive representations, which are obtained by training a model to predict future states using an autoregressive transition model. 
However, when states are high-dimensional data, such as images, predicting the raw, high-dimensional state may require an overly complex model to reconstruct fine-grained details of the states.
Therefore, recent studies have focused on predicting future states in the latent space.
They utilized a Siamese model to encode future states into latent representations and trained the model to maximize the similarity between the predicted future latents and ground truth future latents \cite{cpc, schwarzer2020spr, atc, sgi}.

When maximizing the latent representations' similarity with a Siamese model, there exists a potential pitfall known as \textit{representational collapse}, wherein the subspace of the latent representations collapses into a low-dimensional manifold (i.e., low feature rank) \cite{simsiam, onfeature}. 
Having a sufficiently high feature rank is an important factor for the pretrained model to learn downstream tasks, as the feature rank decides the representation power of the model to adapt to new tasks ~\cite{lyle2022capacity, garrido2022rankme}. 
In the extreme case where the feature rank is 0, the encoder will output the same vector for all input states, making it infeasible to learn the policy for downstream tasks.

To prevent representational collapse, contrastive learning \cite{stdim, curl, atc, cpc} or batch-normalization with stop-gradient operation \cite{schwarzer2020spr, sgi} are commonly employed in unsupervised representation learning.
They learn temporally predictive representations without representational collapse by \textit{repulsing} the representation of different states within the same mini-batch while maximizing the similarity between predictions and future states.
However, this repulsion is known to negatively impact representation quality when there exist relevant states in the mini-batch ~\cite{byol, simsiam, khosla2020supclr}.
This problem can be amplified in the unsupervised state representation for RL since each mini-batch generally consists of a batch of consecutive, relevant states.





In response, we introduce a simple temporally predictive representation learning framework for RL (SimTPR), which prevents representational collapse without repulsing the representations of different states in the mini-batch.
Instead, motivated from \cite{cogswell2015reducingdecorr, onfeature, barlow}, we directly maximize the feature rank of the latent manifold by standardizing the cross-correlation matrix between the predictions and the targets in the latent space.
By enforcing the off-diagonal entries of the cross-correlation matrix to 0, this objective makes features to be independent and increases the feature rank of the latent representation's manifold.
An overview of SimTPR is illustrated in Figure \ref{fig:intro}.



Through extensive experiments, we show that SimTPR achieved state-of-the-art performance in unsupervised representation learning on the Atari benchmark.
Following the evaluation protocol from  \cite{sgi, lightweight_probe}, we finetuned an MLP-based policy layer for 100k steps on top of the frozen encoder. 
When pretrained from a state dataset,  SimTPR achieved a human-normalized IQM score of 0.451, representing a 10\% improvement over the previous best, unsupervised state representation learning method, ATC \cite{atc}.
Furthermore, through empirical studies, we discovered that feature rank is an important factor that affects the performance of downstream tasks.
Also, we found that increasing the feature rank with the repulsive methods can harm the representation quality as it pushes away the relevant states in the mini-batch.

\section{Related Work}

\subsection{Unsupervised Representation Learning for Reinforcement Learning.}

Unsupervised state representation learning has a long history of research in RL, where a common underlying principle is to learn temporally predictive representations ~\cite{dayan1993seminar, tesauro1995tdlearning, littman2001predictive, singh2003learning}. 
To learn temporally predictive representations, models are generally trained to predict future states using an autoregressive transition model. 
When the state consists of high-dimensional data, such as images, high-capacity models are introduced to precisely predict future state information \cite{doerr2018probabilisticrnn, buesing2018queryrloriginal, hafner2019rssm}.
However, predicting the full-state information may result in representations that contain information that is irrelevant to finding an optimal policy. 
For example, consider a task of maze navigation, where a TV is placed in the maze displaying random images.
Although the TV content is irrelevant to the task, the model aims to learn representations to predict the screen of TV.
To mitigate this issue, recent works have focused on predicting the latent information of future states, in which a Siamese model is adopted to encode the future states into latent representations.
Then, the model is trained to maximize the similarity between the predictions and the future states in the latent space~\cite{cpc, hafner2019dreamer, atc, schwarzer2020spr, sgi}.


However, solely maximizing similarity can result in a representational collapse, in which the subspace of the latent representations collapses into a low-dimensional manifold or even to a constant.
To prevent representational collapse, recent studies have employed the contrastive learning \cite{stdim, curl, cpc, atc} or architectural designs including the use of batch-normalization and stop-gradient operation \cite{schwarzer2020spr, sgi}.
They prevent representational collapse by repulsing the representations within the same mini-batch.
While effective in preventing representational collapse, this repulsion has been noted to have a risk of pushing away semantically relevant states \cite{byol, simsiam, khosla2020supclr}. 
This problem can be further amplified in state representation learning, where each mini-batch generally consists of a batch of consecutive states \cite{schwarzer2020spr}.


In light of these limitations, we propose an alternative objective, referred to as the \textit{feature decorrelation} loss.  
The feature decorrelation loss aims to mitigate representational collapse by directly maximizing the dimensions of the latent space's manifold, without pushing away representations within a given mini-batch.

\subsection{Unsupervised Visual Representation Learning.}

Unsupervised visual representation learning aims to learn rich representations from high-dimensional, unlabeled images.
Among these approaches, invariance learning is the state-of-the-art representation learning method ~\cite{simclr, he2020moco, byol, simsiam, caron2020swav}.
Invariance learning involves maximizing the similarity between representations of two differently augmented views of the same image in the latent space.
However, solely maximizing the similarity may result in a representational collapse \cite{onfeature, wang2021understandingcont, jing2021understandingcollapse}.


To avoid representational collapse, contrastive learning \cite{simclr, he2020moco} or architectural variants with batch normalization and stop-gradient operation \cite{byol, batchnorm_byol, simsiam} have been proposed.
They prevent the collapse by repulsing the representations of different images in the mini-batch.
Recently, another line of research has explored the use of feature decorrelation objectives in unsupervised visual representation learning~\cite{pica, barlow, onfeature, vicreg}.
This approach aims to maximize the dimensionality of the latent manifold by standardizing the covariance matrix of the representations in the mini-batch ~\cite{huang2018dbn, kessy2018whitening}.
This alternative approach has achieved competitive performance, compared to state-of-the-art methods.

\section{Method}
\label{body:Method}
\begin{figure}[h]
\begin{center}
\includegraphics[width=0.94\linewidth]{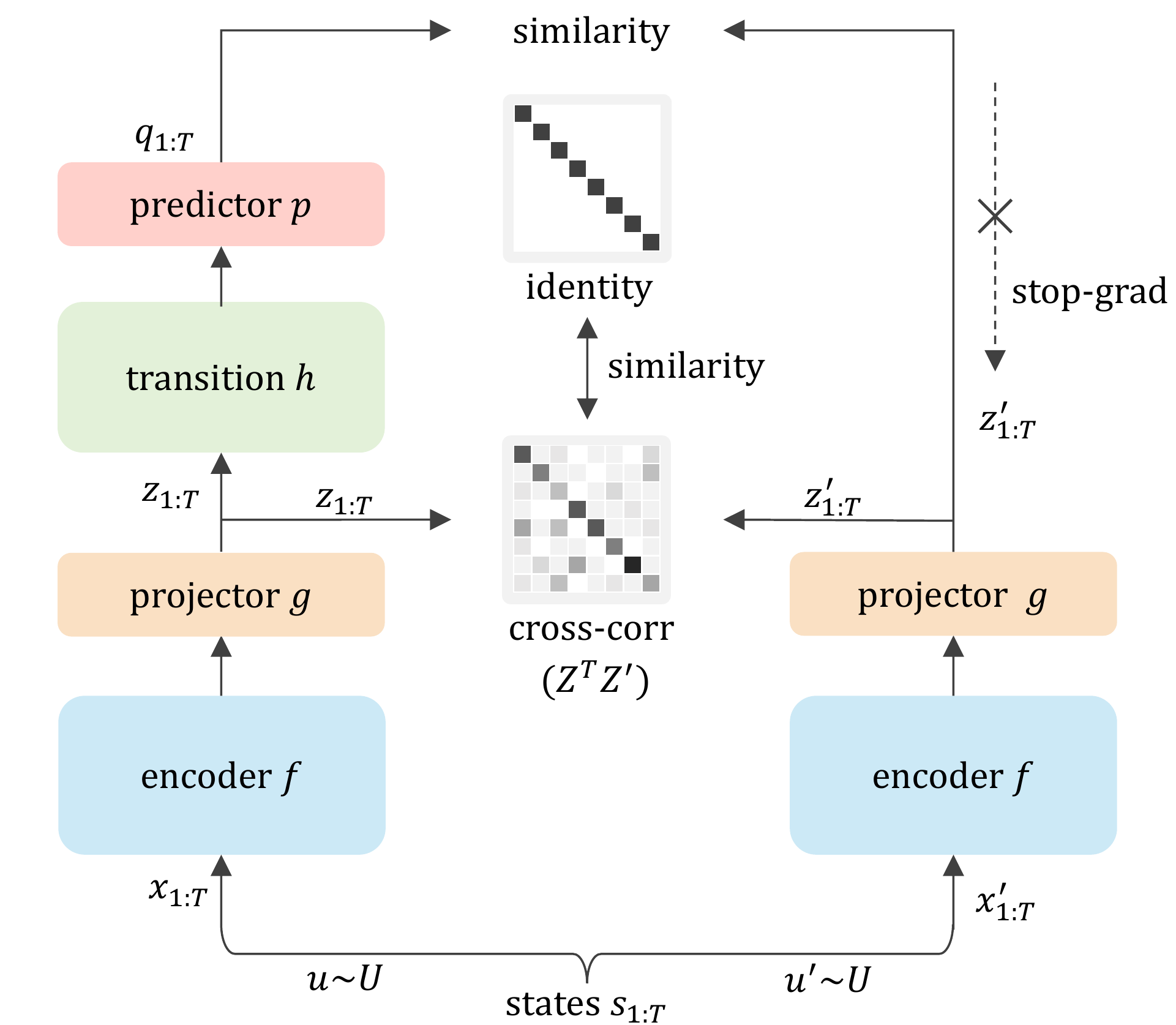}
\end{center}
\vspace{-2mm}
\caption{\textbf{SimTPR architecture.} Two augmented views of a sequence of states are processed with the same encoder $f$ and the same projection MLP $g$. Then a transition model $h$ and prediction MLP $p$ are applied on one side to predict the future states in latent space. While the model maximizes the similarity, it simultaneously standardizes the cross-correlation matrix to prevent collapse.}
\vspace{-2mm}
\label{fig:framework}
\end{figure}

As illustrated in Figure \ref{fig:framework}, SimTPR aims to learn temporally predictive representations by causally predicting the future states within the latent space.
To achieve this, SimTPR utilizes an autoregressive transition model to predict future states and maximizes the similarity between the predictions and the future states in the latent space.
Then, SimTPR integrates a feature decorrelation loss that prevents representational collapse by standardizing the cross-correlation matrix of the representations in the latent space. 


\noindent
\textbf{Pretraining from State.} Given a dataset of states, $\mathcal{S}$, we uniformly sample a sequence of $T$ states $s_{1:T} \sim \mathcal{S}$. 
Then, we sample two different augmentation functions, $u$ and $u'$, from augmentation distribution $\mathcal{U}$.
By applying augmentations, $u$ and $u'$, to each sequence of states, $s_{1:T}$, we generate two different views, $x_{1:T} \triangleq u(s_{1:T})$ and $x'_{1:T} \triangleq u'(s_{1:T})$. 


Two augmented views, $x_{1:T}$ and $x'_{1:T}$, are then processed by four different model components as follows:

\begin{itemize}[noitemsep, topsep=0pt]
    \item A encoder, $f$, which encodes representation vectors from augmented states.  Our framework allows various choices of the network without any constraints. 
    Following previous work \cite{sgi}, we use a 36-layer convolutional network as an encoder, which is the modified version of EfficientNet \cite{tan2019efficientnet}.
    \item  A projector, $g$, which maps encoded representations to the $d$-dimensional latent space. 
    Following \cite{simclr, sgi}, we use a simple one-hidden layer MLP with ReLU activation.
    For each augmented view, we obtain $z_{1:T} = g(f(x_{1:T}))$ and $z'_{1:T} = g(f(x'_{1:T}))$ where $z_{1:T}, z'_{1:T} \in \mathbb{R}^{T \times d}$.
    \item A transition model, $h$, which maps states' latent representations to future time step latent representations. 
    We use a causal transformer \cite{radford2018gpt} to model the transition dynamics.
    
    \item A predictor, $p$, that predicts the representations of the future states in the $d$-dimensional latent space. We use a one-hidden layer MLP with ReLU activation. By applying the transition model followed by the predictor, we obtain  $q_{1:T} = p(h(z_{1:T}))$ where $q_{1:T} \in \mathbb{R}^{T \times d}$.
\end{itemize}

Now, we define our loss function, which composes the \textit{similarity} and \textit{feature decorrelation} loss.

In training, the inputs are processed in batches of $N$ sequences.
We denote the batch of projected representations as 
$Z\hspace{-1mm}=\hspace{-1mm}[z_{1,1:T},..., z_{N, 1:T}]$, $Z'\hspace{-1mm}=\hspace{-1mm}[z'_{1, 1:T}, ..., z'_{N, 1:T}]$, the predicted representations of $Z$ and $Z'$ as $Q\hspace{-1mm}=\hspace{-1mm}[q_{1, 1:T}, ..., q_{N, 1:T}]$, and $Q'\hspace{-1mm}=\hspace{-1mm}[q'_{1, 1:T}, ..., q'_{N, 1:T}]$ respectively, where $Z, Z', Q, Q'\in\mathbb{R}^{N \times T \times d}$. 
We then $\ell_2$-normalize each state representation from $Z, Z',Q, Q'$ to $\bar{Z}, \bar{Z}',\bar{Q}, \bar{Q'}$, where 
$\bar{z}_{i,j}  \hspace{-1mm}=\hspace{-1mm}  z_{i,j}  /  \| z_{i,j}\|_2$, 
$\bar{z}'_{i,j} \hspace{-1mm}=\hspace{-1mm}  z'_{i,j} /  \| z'_{i,j}\|_2$, 
$\bar{q}_{i,j}  \hspace{-1mm}=\hspace{-1mm}  q_{i,j}  /  \|q_{i,j}\|_2$ and 
$\bar{q}'_{i,j} \hspace{-1mm}=\hspace{-1mm}  q'_{i,j} /  \| q'_{i,j}\|_2$.

First, following \cite{byol, simsiam}, we maximize the \textit{similarity} between the predictions and the targets by minimizing their mean squared error as:
\begin{equation}
    \mathcal{D}(\bar{Q},\bar{Z}') \hspace{-1mm}=\hspace{-1mm}
        -\frac{1}{N(T\hspace{-1mm}-\hspace{-1mm}k)} \sum_{n=1}^N  \sum_{t=1}^{T-k} 
        \|\bar{q}_{n,t} - \bar{z}'_{n,t+k} \|_2^2,
\label{eq:cosine distance}
\end{equation}
where $k$ denotes the number of future steps to associate the predictions.
For the sake of simplicity, we set $k=1$.

This similarity loss can be also interpreted as maximizing the cosine-similarity between the un-normalized representations: $Q$ and $Z$ \cite{simsiam, byol}.

Then, we symmetrized a loss function as \cite{byol}:
\begin{equation}
    \mathcal{L}_{\text{sim}} \hspace{-1mm}=\hspace{-1mm} 
    \frac{1}{2} \mathcal{D}(\bar{Q},\text{sg}(\bar{Z}')) + \frac{1}{2} \mathcal{D}(\bar{Q}',\text{sg}(\bar{Z})), 
\label{eq:similarity loss}
\end{equation}
where $\text{sg}(\cdot)$ is stop-gradient operation.

Second, we define a feature decorrelation loss, $\mathcal{L}_{\text{decorr}}$, which maximizes the dimensions of the latent representation's manifold by standardizing the cross-correlation matrix. 

The cross-correlation matrix, $\mathcal{C}$, between $\bar{Z}$ and $\bar{Z}'$ is:
\begin{equation}
    \mathcal{C}(\bar{Z}, \bar{Z}')_{i,j} \hspace{-1mm}=\hspace{-1mm} \frac{\sum\limits_{n=1}^N \sum\limits_{t=1}^T \bar{z}_{n,t,i} \cdot \bar{z}'_{n,t,j}}{\sqrt{\sum\limits_{n=1}^N \sum\limits_{t=1}^T (\bar{z}_{n,t,i})^2} \sqrt{\sum\limits_{n=1}^N \sum\limits_{t=1}^T (\bar{z}'_{n,t,j})^2}}
\label{eq:cross-correlation}
\end{equation}
where $i$ and $j$ indicate the dimension of the vectors.

Then, following \cite{barlow}, we standardize the cross-correlation matrix as:
\begin{equation}
    \mathcal{L}_{\text{decorr}} = 
    \sum_i (1 - \mathcal{C}_{ii})^2 + \lambda_o \sum_i \sum_{j \neq i} \mathcal{C}_{ij}^2
\label{eq:decorrelation loss}
\end{equation}
where $\lambda_o$ controls the importance between two terms. 

Intuitively, the first term in equation \ref{eq:decorrelation loss} generates low-variance features by enforcing the on-diagonal terms to be 1. 
The second term in equation \ref{eq:decorrelation loss} generates features that are independent of each other by enforcing the off-diagonal terms to be 0.  
While the second term is a crucial component to prevent the representational collapse, the first term is also important to induce the encoded representations to be not dependent on a few, high-variance features.

The overall loss function, $\mathcal{L}_{\text{state}}$, is a weighted average of the $\mathcal{L}_{\text{sim}}$ and $\mathcal{L}_{\text{decorr}}$, as:
\begin{equation}
    \mathcal{L} = 
    \mathcal{L}_{\text{sim}} + 
    \lambda_d \mathcal{L}_{\text{decorr}},
\label{eq:video loss}
\end{equation}
where $\lambda_d$ controls the importance between two terms.
The pseudo-code is described in Algorithm \ref{algo:SimTPR_video}.

\noindent
\\
\textbf{Pretraining from Demonstration.}  
In the case where our dataset contains action labels, we can simply extend our framework to further predict the action labels on top of the future state prediction task.
Here, we slightly modify the transition model's input to incorporate the action information and initialize another predictor model to predict the action labels.
We refer to this action predictor as $r$ and use a one-hidden layer MLP with ReLU activation that outputs the number of actions, $n_a$. The rest of the components remain identical to the state pretraining setup.  

Given a dataset of states, $\mathcal{S}$, and actions, $\mathcal{A}$, we sample a sequence of states $s_{1:T} \sim \mathcal{S}$, with their corresponding actions 
$a_{1:T} \sim \mathcal{A}$ .
The actions are then linearly embedded into $d$-dimensional vectors, represented as $y_{1:T} \in \mathbb{R}^{T \times d}$.

Then, we construct the trajectory representation as $ \tau = [z_1, y_1, ..., z_T, y_T] \in \mathbb{R}^{2T \times d}$.
By applying the transition model $h$, we obtain the context representations, $c=h(\tau)$.
This context representation is then decomposed into state and action representations which are processed by the state predictor $p$, and the action predictor $r$ respectively as, $q_{1:T} = p(\{c_i\}_{i=2,4,...,2T})$ and $l_{1:T} = r(\{c_i\}_{i=1,3,...,2T-1})$.

In training, the inputs are processed in batches of $N$ sequences.
We denote the batch of action prediction logits as $L = [l_{1,1:T},..., l_{N, 1:T}]$ and the action labels as $A = [a_{1,1:T},..., a_{N, 1:T}]$, where $L \in \mathbb{R}^{N \times T \times n_a}$ and $A \in \mathbb{R}^{N \times T}$. 

Then, we define the action prediction loss as:
\begin{equation}
    \mathcal{L}_{\text{act}} = -\frac{1}{NT} \sum_{n=1}^N  \sum_{t=1}^{T} \log(\frac{\exp(l_{n,t,a_{\scaleto{n,t}{3pt}}})}{\sum_{i=1}^{n_a} \exp(l_{n,t,i})}) 
\label{eq:act loss}
\end{equation}
which minimizes the negative log-likelihood of the action for each state. 

The overall loss function, $\mathcal{L}$, is a weighted average of the $\mathcal{L}_{\text{sim}}$, $\mathcal{L}_{\text{decorr}}$, and  $\mathcal{L}_{\text{act}}$, as:
\begin{equation}
    \mathcal{L} = 
    \mathcal{L}_{\text{sim}} + 
    \lambda_d \mathcal{L}_{\text{decorr}} +
    \lambda_a \mathcal{L}_{\text{act}},
\label{eq:demonstration loss}
\end{equation}
where $\lambda_a$ controls the importance of the action prediction loss.
The pseudo-code is described in Algorithm \ref{algo:SimTPR_demon}.


\section{Experiments}


\subsection{Experimental Setup}

Here, we present the experimental setups. 
A detailed explanation for each setup is described in Appendix \ref{appendix:impl_details}.

In this study, we evaluate our method using the Atari100k benchmark \cite{bellemare2013arcade}. This benchmark is widely used to measure the sample efficiency of Reinforcement Learning (RL) algorithms \cite{drq, schwarzer2020spr, curl}.
Following \cite{sgi, lightweight_probe}, we pretrain the models on 26 different games in the Atari environment and finetune each model for 100k steps of online interactions.

\textbf{Pre-training.} We follow the pretraining protocol from \cite{sgi, lightweight_probe}, where we use the publicly-available DQN replay dataset \cite{agarwal2020optimistic}.
It contains the training logs of a DQN agent for 50 million steps. 
We select 1.5 million frames from the 3.5 to 5 million steps of the replay dataset as~\citet{lightweight_probe}.
This selection corresponds to the training logs of a weak, partially trained agent, which reflects the nature of publicly available datasets in internet platforms that generally contain sub-optimal trajectories \cite{baker2022vpt}. 


As our default configuration, we use two simple image augmentations used in \cite{schwarzer2020spr, sgi, lightweight_probe, atc} which are \textit{random shift} followed by an \textit{intensity jittering}.
For the encoder, we use a 30-layer convolutional network from \cite{sgi} and a 2-layer MLP as a projector, which projects to a $512$-dimensional latent space. 
For the transition model, we use two layers of causal transformer block \cite{radford2018gpt} with $512$ hidden dimensions. 
The predictor network is also a 2-layer MLP which maps to a $512$-dimensional latent space.
To balance the weights between the loss functions, we set $\lambda_o=0.005, \lambda_d = 0.01$, and $\lambda_a=1.0$.
For each training step, a training batch consists of $640$ samples with $N=64$ and $T=10$.
We use an AdamW optimizer \cite{loshchilov2017adamw} with a learning rate of $3 \times 10^{-4}$ and a weight decay of $10^{-6}$ and train for 100 epochs. 

To identify representational collapse, we measure the feature rank (Feat.Rank) of the latent representations during the training process. In training, we randomly sample $n$ states and feed them into the model to obtain the projection matrix, $Z \in \mathbb{R}^{n \times d}$. Then, we perform the singular value decomposition to the projection matrix, $Z$, and count the number of singular values of its diagonal matrix that are larger than a constant, $\epsilon$. This estimation measures the dimension of the subspace spanned by the features in $Z$, after removing the highly correlated features \cite{golub1976rank, lyle2022capacity}.
We use $n=1000$ and $\epsilon=0.01$.


\textbf{Linear Probing.} To evaluate the learned representations, we follow the linear evaluation protocol from \cite{lightweight_probe}. After pretraining, we train a linear classifier on top of the frozen encoder and predict the reward or action of the expert.
For the expert dataset, we use the last 100k frames from the DQN replay dataset with a 4:1 train/eval split.

For reward prediction, we simplify the problem to a binary classification task and predict whether a reward has occurred or not in a given state. We report the mean F1 score across environments (Rew F1).
For action prediction, we train a multiclass classifier and report the multiclass F1 score across environments (Act F1).
These scores serve as a proxy to measure whether the agent can effectively learn the policy and the value function of the experts. 

\textbf{Finetuning.} For each environment, we finetune the pretrained model for 100k steps. We follow the protocol from \cite{sgi, lightweight_probe}, where we train an MLP-based Q-learning policy on top of the frozen encoder using the Rainbow algorithm \cite{hessel2018rainbow}.
We do not use any other auxiliary loss during finetuning. 

For each game, we compute the average score of 50 trajectories, evaluated at the end of the training. Then, this score is normalized as HNS=$\frac{\text{agent\_score - random\_score}}{\text{human\_score - random\_score}}$, which measures the relative performance to humans.
To reduce the variance of each run, this score is averaged over 5 random seeds for our empirical studies and 10 random seeds for our main results, including all baselines. 
Following the guidelines of \cite{agarwal2021iqm}, we report a bootstrapped interval for the mean, median, interquartile mean (IQM), and optimality gap (OG) of HNS over 26 games. OG estimates the gap of the average probability to satisfy the HNS by 1.0.


\subsection{Empirical Study}

In this section, we conduct empirical studies to investigate the importance of the feature rank and the potential pitfalls of the repulsive methods in maximizing the feature rank.
We focus on unsupervised representation learning where the encoder is pretrained from the state dataset.

\textbf{Effect of Feature Decorrelation.} First, we analyze the effect of the decorrelation loss by varying the decorrelation strength $\lambda_d$ for SimTPR.  
We fix the remaining hyperparameters to its optimal configuration.

\begin{figure}[h]
\begin{center}
\includegraphics[width=0.49 \textwidth]{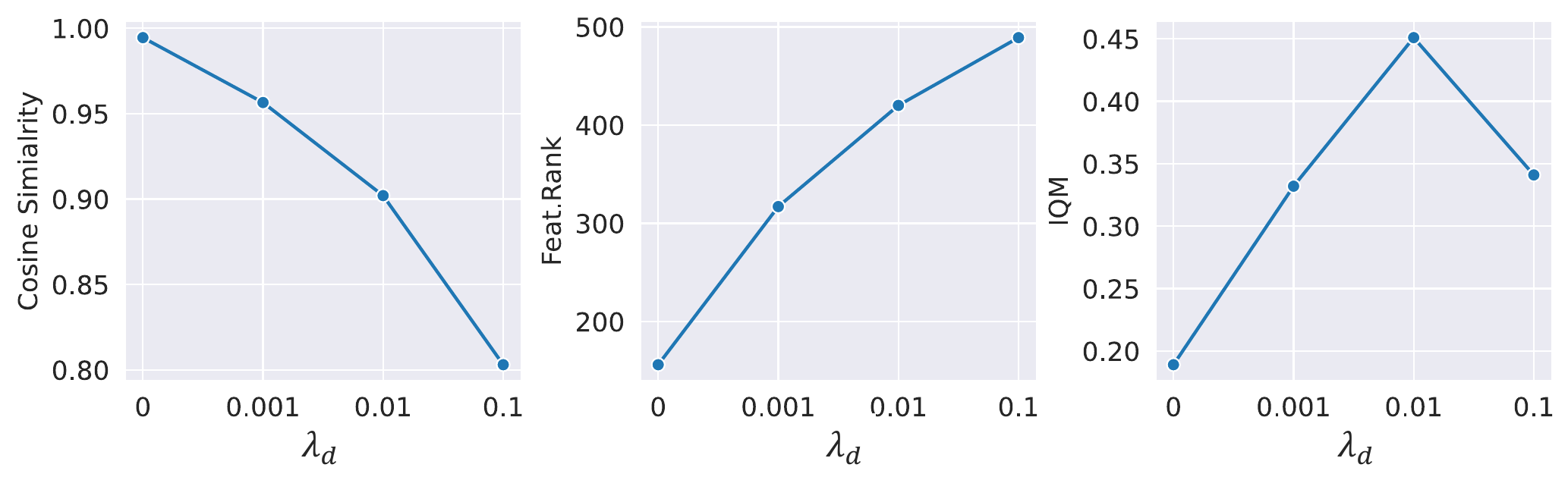}
\end{center}
\vspace{-4mm}
\caption{\textbf{Effect of decorrelation strength ($\lambda_d$).}
There exists a trade-off between the similarity loss and the decorrelation loss where SimTPR achieved the highest IQM  when $\lambda_d=0.01$.}
\label{fig:decorr_curve}
\vspace{-2mm}
\end{figure}

Figure~\ref{fig:decorr_curve} illustrates the effect of the decorrelation strength on the representation similarity (i.e., the cosine similarity between the predicted states and future states), feature rank, and finetuning performance of SimTPR. 
In Figure~\ref{fig:decorr_curve}, we observed that stronger decorrelation leads to an increase in the feature rank while it leads to a decrease in the predictive similarity.
Among the variants, the SimTPR with a moderate decorrelation strength ($\lambda_d = 0.01$) shows the highest finetuning performance, with an IQM score of 0.451.
These empirical findings reveal that there exists a trade-off between learning temporally predictive representations and feature decorrelation.
Thus, it is important to balance these two terms to achieve good finetuning performance.

\begin{figure}[h]
\begin{center}
\includegraphics[width=0.4 \textwidth]{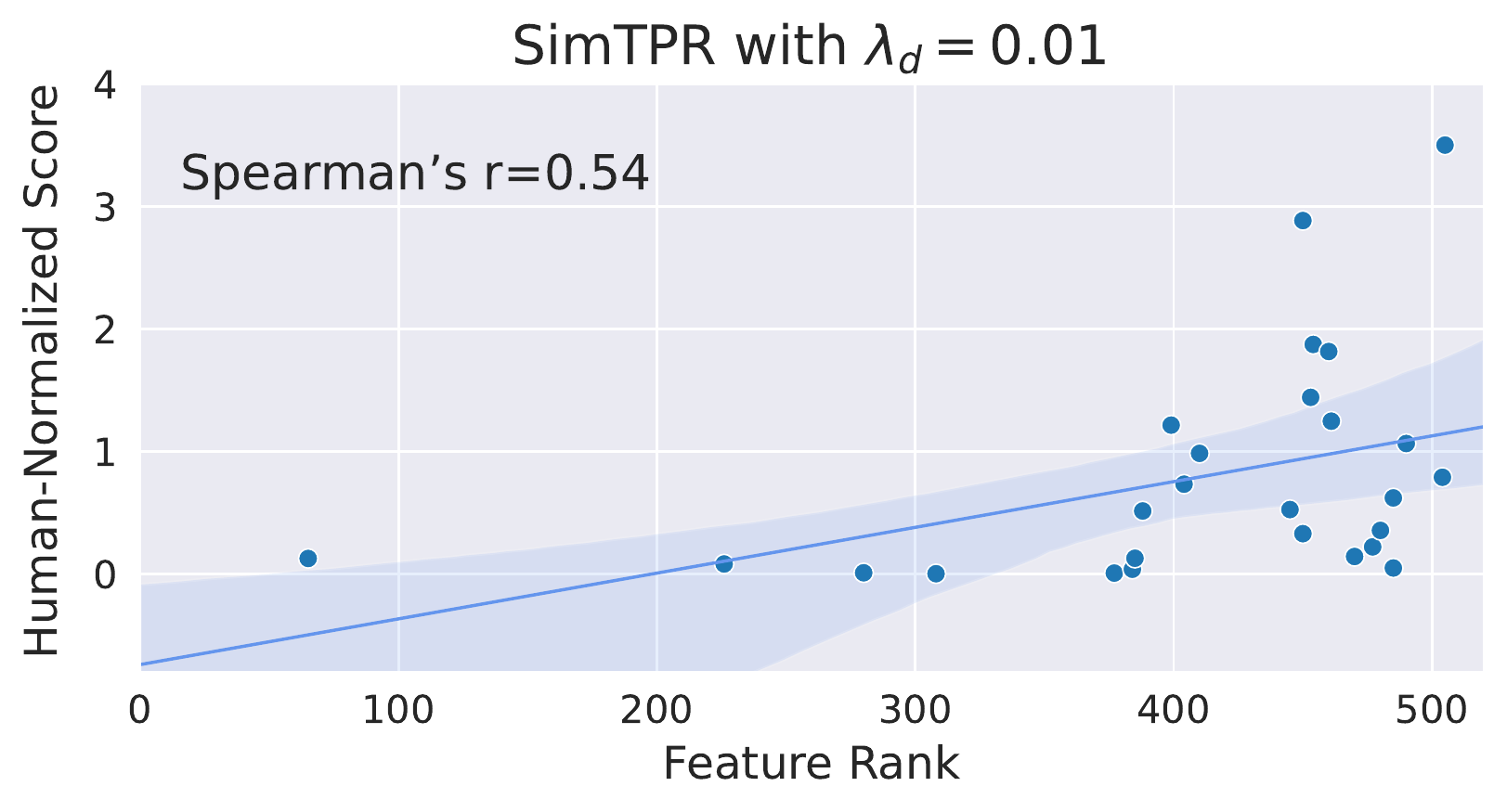}
\end{center}
\vspace{-4mm}
\caption{\textbf{Relationship between feature rank and finetuning performance.}
The figure presents the human-normalized score of each game, plotted against the corresponding feature rank of the pretrained model. 
While the feature rank is not the sole determinant of agent performance, 
there exists a positive correlation between feature rank and human-normalized score.}
\label{fig:rank_per_game}
\end{figure}

In Figure~\ref{fig:rank_per_game}, we illustrate the human-normalized score for each individual Atari game along with their respective feature rank of the pretrained encoders.
Here, we use SimTPR with decorrelation strength of $\lambda_d = 0.01$,
The scatterplot in the figure illustrates that there exists a clear correlation between feature rank and human-normalized scores, as quantified by Spearman's correlation coefficient of $r = 0.54$. 
While there exist several games with low fine-tuning performance with high feature rank, all the games with low feature rank ($\leq$ 300) show low fine-tuning performance.

In summary, our findings indicate that although there are numerous factors (e.g., credit assignments, learning algorithm, optimizer) that affect the learning process of the RL agent,
having a sufficiently high feature rank is an important factor to attain good finetuning performance.

\textbf{Feature Decorrelation vs Repulsion.}  
In this section, we compare the feature decorrelation loss to repulsion-based methods. 
Specifically, we consider two repulsion-based methods: contrastive learning and batch normalization (BN). 
These methods prevent collapse by repulsing the representations of different states in a mini-batch. 
For contrastive learning, we replace the similarity loss $\mathcal{L}_{\text{sim}}$ with the contrastive loss $\mathcal{L}_{\text{cont}}$ as in \cite{simclr} and vary the repulsion temperature from $\{0.05, 0.1, 0.2\}$ to select the best trade-off between similarity maximization and repulsion. 
We use the temperature of $0.1$ which yields the highest finetuning performance.
For BN, we apply the BN layer to the output of the first layer of the projector.

\begin{table}[h]
\vspace{-3mm}
\caption{\textbf{Comparison of collapse prevention methods}. The latent dimension is 512. The grey row indicates our default setup. }
\begin{center}
\resizebox{0.42 \textwidth}{!}{
\begin{tabular}{l ccccc}
\toprule
& $\mathcal{L}_{\text{cont}}$ & batch-norm & $\mathcal{L}_{\text{decorr}}$ & Feat.Rank & IQM $\uparrow$ \\
\midrule \\[-2.4ex]
(a) &  -         & -           &-          & 156     & 0.189  \\
(b) & \checkmark & -           &-          & 443     & 0.331  \\
(c) & -          & \checkmark  &-          & 305     & 0.068  \\
\rowcolor{shadecolor}
(d) & -          & -           &\checkmark & 421     & \textbf{0.451}  \\
(e) & \checkmark & -           &\checkmark & 472     & 0.342  \\
(f) & -          & \checkmark  &\checkmark & 500     & 0.264  \\
\bottomrule
\end{tabular}}
\label{table:decorr_video}
\end{center}
\end{table}

Table \ref{table:decorr_video} reports the finetuning performance for different configurations of collapse prevention methods.
By comparing Table \ref{table:decorr_video}.(a) and (b), 
we observed that using the contrastive loss effectively alleviates representational collapse (Feat.Rank=443), resulting in an improved IQM score of 0.331. 
In Table~\ref{table:decorr_video}.(c), we found that although batch-normalization alleviates the collapse (Feat.Rank=305), it results in a significant drop in finetuning performance, with 0.068 IQM. 
We speculate that applying batch-normalization brings extra difficulties in training the transition model, as it makes the input, $z_{1:T}$, and the target, $z'_{1:T}$, of the transition model to fluctuate depending on the distribution of the mini-batch.
Among the variants, SimTPR with the decorrelation loss shows the best finetuning performance with 0.451 IQM (Table \ref{table:decorr_video}.(d)).
From Table \ref{table:decorr_video}.(e),(f) we observed that further using contrastive loss or batch normalization leads to a higher feature rank but decreases the finetuning performance with an IQM of 0.342 and 0.260 respectively.

Although both contrastive and feature decorrelation loss has shown to effectively prevent representation collapse (i.e., high feature rank), we observed a large gap of IQM scores between these methods (Table \ref{table:decorr_video}.(b),(d)). 
We hypothesize that this discrepancy stems from the repulsive force of the contrastive loss which may even separate the relevant states in the mini-batch.
To validate our hypothesis, we conduct an in-depth comparison of the pretrained representations of the contrastive loss to feature decorrelation loss.



\begin{figure}[h]
\begin{center}
\includegraphics[width=0.49 \textwidth]{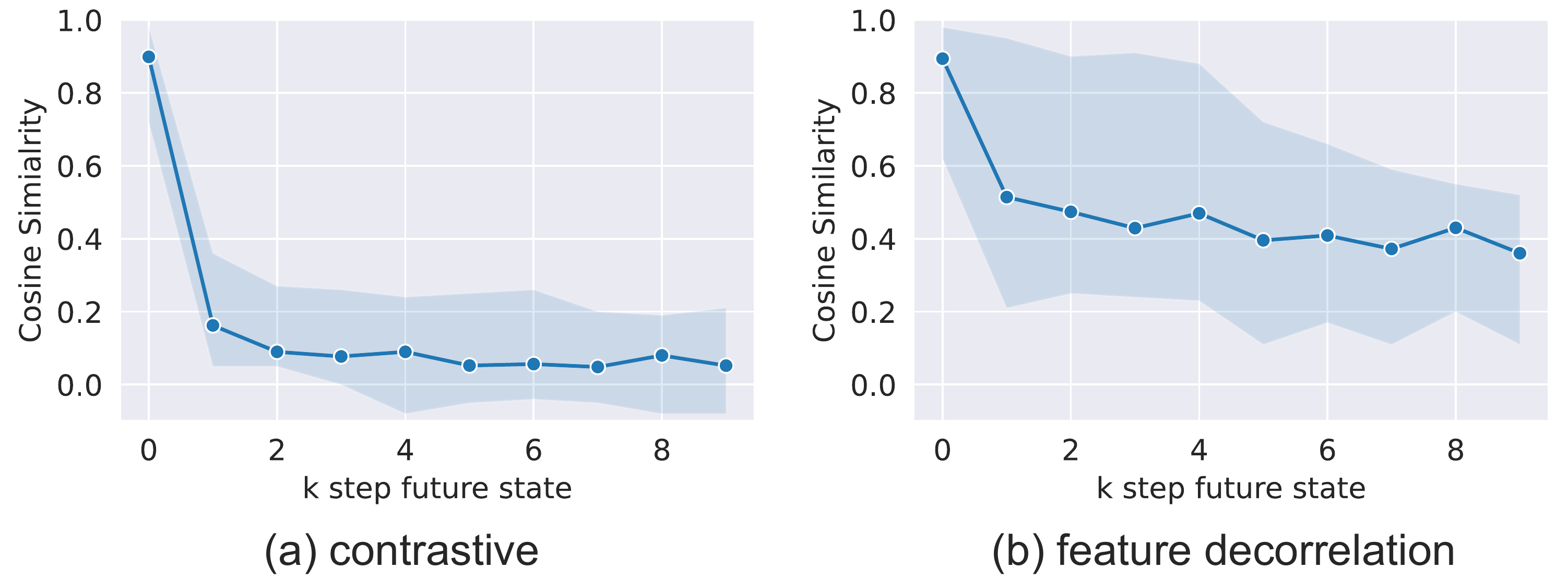}
\end{center}
\vspace{-3mm}
\caption{\textbf{Cosine Similarity between state representations}. 
The figure presents the average cosine similarity across 26 Atari games.}
\label{fig:cosine_sim}
\end{figure}

\begin{table*}[t]
\label{table: main_result}
\caption{\textbf{Finetuning on Atari-100k benchmark}. In pre-training, all methods are trained for 100 epochs with DQN replay dataset. 
Then, we finetuned a 2-layer Q-learning policy on top of the frozen encoder for 10 random seeds. We report 95\% Confidence Intervals (CI) of IQM, Median, Mean, and Optimality Gap scores where CIs are estimated using the percentile bootstrap with stratified sampling.
All competitors except SGI \cite{sgi} and BarlowBalance \cite{lightweight_probe} based on our reproduction.}
\begin{center}
\resizebox{0.95 \textwidth}{!}{
\begin{tabular}{lcccccc}
\toprule
Method &Act F1 $\uparrow$ & Rew F1 $\uparrow$
& IQM $\uparrow$   & Median $\uparrow$ & Mean $\uparrow$ & Optimality Gap $\downarrow$ \\
\midrule \\[-2.5ex]

\textit{No Pretraining} \\[0.5ex]
DrQ \cite{drq}    
& - 
& - 
& 0.161 \scriptsize{(0.149, 0.174)}  & 0.157 \scriptsize{(0.129, 0.184)} 
& 0.308 \scriptsize{(0.285, 0.332)}  & 0.724 \scriptsize{(0.710, 0.737)}    \\
\midrule \\[-2.5ex]

\textit{Pretrained from State} \\[0.5ex]
VAE \cite{kingma2014vae} 
& 22.1    & 56.8        
& 0.266 \scriptsize{(0.236, 0.301)}   & 0.266 \scriptsize{(0.198, 0.293)}         
& 0.556 \scriptsize{(0.498, 0.620)}   & 0.623 \scriptsize{(0.600, 0.645)}           \\
BarlowTwins \cite{barlow} 
& 21.3    & 58.0        
& 0.224 \scriptsize{(0.203, 0.247)}   & 0.249 \scriptsize{(0.204, 0.289)} 
& 0.448 \scriptsize{(0.410, 0.488)}   & 0.662 \scriptsize{(0.644, 0.680)}  \\
CURL \cite{curl} 
& 15.5    & 52.5        
& 0.247 \scriptsize{(0.222, 0.273)}   & 0.208 \scriptsize{(0.185, 0.264)} 
& 0.426 \scriptsize{(0.400, 0.452)}   & 0.654 \scriptsize{(0.636, 0.673)}  \\
RSSM \cite{hafner2019rssm} 
& 23.6    & 61.1        
& 0.302  \scriptsize{(0.255, 0.348)}  & 0.235 \scriptsize{(0.199, 0.288)}        
& 0.707  \scriptsize{(0.611, 0.811)}  & 0.595 \scriptsize{(0.571, 0.620)}  \\
ATC  \cite{atc}
& 25.8    & 65.6        
& 0.353 \scriptsize{(0.325, 0.384)}   & 0.376 \scriptsize{(0.262, 0.409)} 
& 0.647 \scriptsize{(0.585, 0.714)}   & 0.570 \scriptsize{(0.553, 0.586)}  \\
SimTPR (ours)
& \textbf{25.9}  & \textbf{67.7}  
& \textbf{0.451} \scriptsize{(0.410, 0.494)}   & \textbf{0.434} \scriptsize{(0.317, 0.507)} 
& \textbf{0.773} \scriptsize{(0.707, 0.837)}   & \textbf{0.522} \scriptsize{(0.503, 0.541)}  \\
\midrule \\[-2.5ex]

\textit{Pretrained from Demonstration} \\[0.5ex]
BC
& \textbf{27.1}    & 66.2       
& 0.413 \scriptsize{(0.376, 0.450)}   & 0.344 \scriptsize{(0.289, 0.402)} 
& 0.705 \scriptsize{(0.660, 0.751)}   & 0.536 \scriptsize{(0.517, 0.555)}  \\

IDM
& 26.5     & 64.7      
& 0.343  \scriptsize{(0.308, 0.380)}   & 0.279 \scriptsize{(0.237, 0.337)}        
& 0.564  \scriptsize{(0.520, 0.614)}   & 0.579 \scriptsize{(0.556, 0.601)}  \\


SGI \cite{sgi}
& 26.8    & 64.0        
& 0.380 \scriptsize{(0.329, 0.436)}   & 0.490 \scriptsize{(0.355, 0.573)} 
& 0.751 \scriptsize{(0.678, 0.828)}   & 0.557 \scriptsize{(0.530, 0.587)}  \\

BarlowBalance \cite{lightweight_probe}
& 22.7    & 61.3        
& 0.338 \scriptsize{(0.296, 0.382)}   & 0.201 \scriptsize{(0.131, 0.316)}      
& \textbf{1.089} \scriptsize{(0.983, 1.203)}   & 0.588 \scriptsize{(0.569, 0.607)}           \\

SimTPR (ours)
& 26.9    & \textbf{71.2}   
& \textbf{0.500} \scriptsize{(0.464, 0.537)}   & \textbf{0.515} \scriptsize{(0.381, 0.572)} 
& 0.757 \scriptsize{(0.715, 0.800)}   & \textbf{0.493} \scriptsize{(0.476, 0.511)}  \\

\bottomrule
\end{tabular}}
\end{center}
\vskip -0.1in
\end{table*}
\label{table:main_result}

First, we investigate the cosine similarity between the latent representation, $z_t$, and the $k$-step distant latent representations, $z_{t+k}$, of the pretrained encoder.
In Figure \ref{fig:cosine_sim}, we plot the average similarity of the latent representations w.r.t $k$.
As the distance $k$ increases, we observed a drastic decrement of similarity for contrastive loss whereas the feature decorrelation loss was able to maintain a relatively higher level of similarity.

\begin{figure}[h]
\begin{center}
\includegraphics[width=0.45 \textwidth]{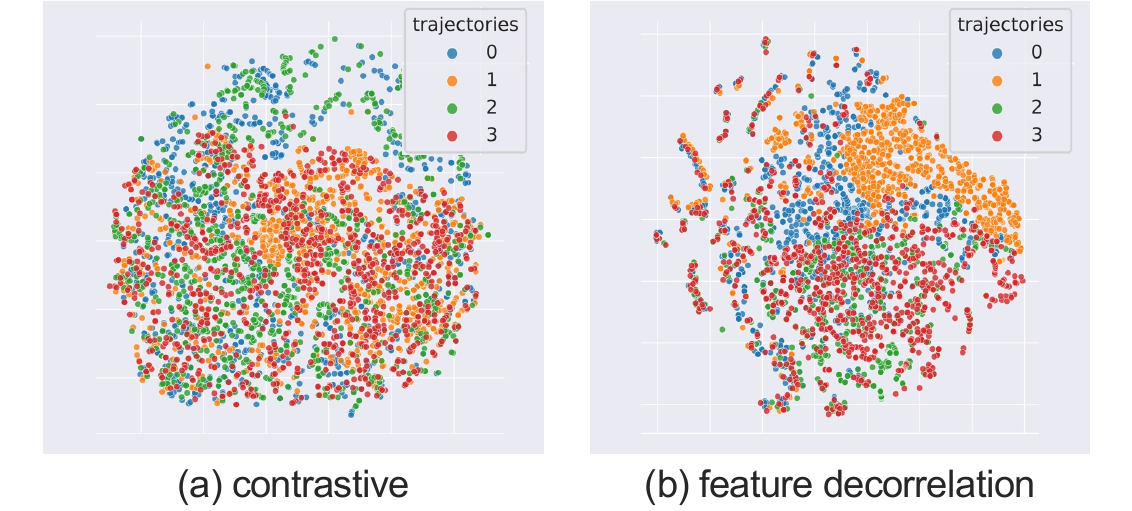}
\end{center}
\vspace{-2mm}
\caption{\textbf{t-SNE visualization.}
We visualize the t-SNE plot of the latent representations on the Pong environment pretrained with (a) contrastive loss and (b) feature decorrelation loss. We use 4 distinct trajectories, each with a length of $T=1000$.
}
\label{fig:tsne}
\end{figure}


In Figure \ref{fig:tsne}, we further visualize the t-SNE plot of the latent representations.
We use the Pong environment and visualize the four distinct trajectories with 1000 steps for each trajectory.
While the latent representations of the contrastive loss were relatively spread out regardless of the trajectories,
the latent representations of the feature decorrelation loss were clustered based on the trajectories.

In summary, we found that although repulsive methods can effectively prevent representational collapse,
they push away the representations of the consecutive, relevant states.
We believe that this repulsion between the consecutive states is a key factor that brings the decrement of the performance for the contrastive loss to the decorrelation loss. 

\subsection{Main Result}

In Table \ref{table:main_result}, we compare SimTPR with state-of-the-art unsupervised representation learning methods in RL. 
To ensure a fair comparison, we fix the encoder, latent dimension, augmentation, pretraining data, and the number of pretraining steps across all methods. 
The results for SGI and BarlowBalance were obtained by running their official code, whereas the results for the remaining baselines are based on our reproduced results.
When reproducing baselines, we strictly followed the training recipes outlined in the original papers.
Further details are described in Appendix \ref{appendix:baselines}.

When pretrained from state datasets, we observed that pretraining consistently led to improved performance compared to learning from scratch. In addition, we discover that methods that learn temporally predictive representations (i.e., RSSM, ATC, SimTPR) achieved superior performance compared to methods that do not learn any temporal dynamics (i.e., VAE, BarlowTwins, CURL).
Among these methods, SimTPR achieved the highest scores in all metrics, with an improvement of 10\% in IQM (0.451) over the previous best method, ATC (0.353).

Training with demonstration datasets generally leads to enhanced finetuning performance compared to the methods with pretraining from states, since the action labels provide additional information to learn good representations.
By leveraging action labels, the finetuning performance of SimTPR increases to an IQM score of 0.500, which is the best result among the demonstration pretraining methods.
Although BarlowBalance achieved the highest mean score, this is due to a few games with exceptionally high human normalized scores (HNS of 6.7 in the game Up'n Down).

\section{Discussion}

Here, we present ablations and discussions for SimTPR to give an intuition of its behavior and performance.
In this section, we focus on pretraining SimTPR from states. 

\subsection{Robustness to Batch Size and Latent Dimension}

First, we investigate the robustness of SimTPR w.r.t different batch sizes and latent dimensions. We maintained the other hyperparameters identical to their optimal configurations.

\begin{figure}[h]
\begin{center}
\includegraphics[width=0.49 \textwidth]{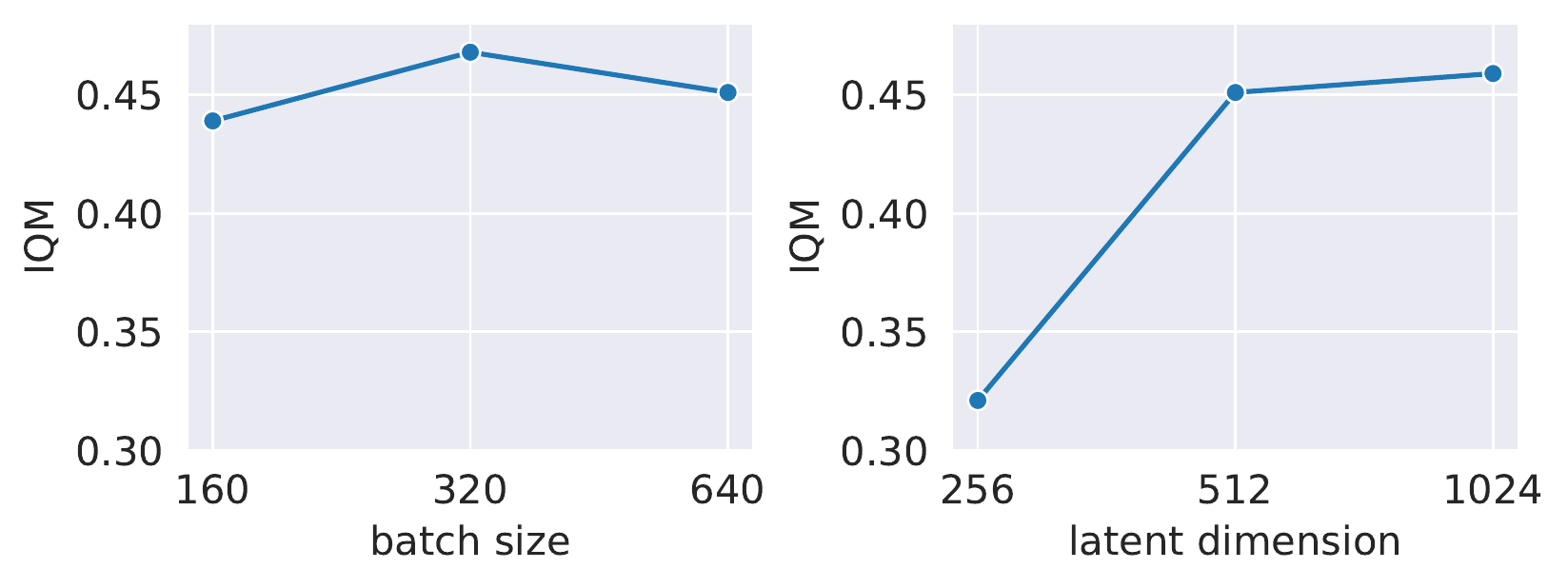}
\end{center}
\vspace{-2mm}
\caption{\textbf{Effect of batch size and latent dimension.} SimTPR was robust to the changes of batch size and showed a positive correlation between the latent dimension and IQM.
}
\label{fig:ablation1}
\end{figure}

Figure \ref{fig:ablation1} displays SimTPR's performance with varying batch sizes and latent dimensions. Our observations indicate that SimTPR's performance remains consistent across different batch sizes, underscoring its robustness. Moreover, we observed a positive correlation between the size of the latent dimension and the fine-tuning performance, as quantified by the IQM. This correlation implies that increasing the latent dimension could potentially improve the model's performance by facilitating more expressive representations.

Interestingly, these observations align with the results from \cite{barlow}, who utilized a feature decorrelation loss in vision tasks.
They report that the feature decorrelation loss remains robust to changes in batch size while performance improves as the latent dimension increases.

\subsection{Transition Model: Causal vs Non-Causal}


Here, we discuss the training strategy and architecture of the transition model. 
For the training strategy, we compare the performance of causal \cite{radford2018gpt} and non-causal \cite{kenton2019bert} transition models, using the transformer's encoder and decoder respectively. 
In training the non-causal model, we randomly masked input representations and reconstruct the masked representations.
For the model architecture, we compare the performance of the transformer to the simple recurrent transition model which consists of the 2-layers of GRU block \cite{cho2014gru}.

\begin{table}[h]
\vspace{-2mm}
\caption{\textbf{Comparison of the transition model}. A non-causal and causal transformer uses the encoder and decoder of transformer architecture, respectively. The grey row indicates the default setup.}
\begin{center}
\resizebox{0.48 \textwidth}{!}{
\begin{tabular}{lcc cccc}
\toprule
& strategy & architecture & mask ratio & Feat.Rank & IQM $\uparrow$   \\
\midrule \\[-2.5ex]

(a) & \multirow{3}{*}{non-causal} & \multirow{3}{*}{transformer} & 0.3       & 435  & 0.270 \\
(b) &                             &                              & 0.5       & 478  & 0.385  \\
(c) &                             &                              & 0.7       & 452  & 0.367  \\
\midrule \\[-3.0ex]
\rowcolor{shadecolor}
(d) & causal                      & transformer                    & -         & 421   &  \textbf{0.451}  \\
(e) & causal                      & gru                          & -         & 438   &  0.387  \\
\bottomrule
\end{tabular}}
\label{table:transition}
\end{center}
\end{table}

In Table \ref{table:transition}.(a), we observed that the non-causal variant with a 30\% mask ratio shows the lowest IQM of 0.270.
We speculate that with a low masking ratio, the model may easily reconstruct the masked representation by interpolating the representations of the past and future. 
Although the variants with higher mask ratios showing the improved performance (Table \ref{table:transition}.(b) and (c)),  they still performed worse than the causal model in Table~\ref{table:transition}.(d).
Comparing Table \ref{table:transition}.(e) to (d), we found that a simple recurrent model can learn temporally predictive representations with an IQM score of 0.384, but using a powerful transition model brings further benefits.

\subsection{Finetuning matters}


Throughout this paper, we have focused on evaluating models that were pretrained, followed by finetuning a policy on top of the frozen encoder. 
However, in this section, we broaden our scope to consider other finetuning strategies that can potentially enhance the finetuning performance.


One crucial finding from recent studies is that the loss of plasticity significantly contributes to the sample inefficiency of RL \cite{nikishin2022primacy, cetin2022stabilizing, lyle2023understanding_plasticity}. The term \textit{loss of plasticity} describes the phenomenon where an agent overfits to initial interactions and subsequently struggles to adapt to new data. We find this issue to be especially pronounced in the pretrain-then-finetune paradigm. This is because pretrained models, with their inherent capability to rapidly adapt, are more susceptible to overfitting to earlier interactions.

\begin{table}[b]
\vspace{-2mm}
\caption{\textbf{Comparison of finetuning strategy}. For the reset, we periodically reset the policy layer for every 40,000 update steps.}
\begin{center}
\resizebox{0.39 \textwidth}{!}{
\begin{tabular}{lc lcc}
\toprule
& pretrain
& encoder
& reset policy
& IQM $\uparrow$ \\
\midrule \\[-2.5ex]
(a)
& -
& Frozen
& -        
& 0.161    
\\

\midrule \\[-2.5ex]

(b)
& \multirow{3}{*}{\checkmark} 
& Frozen
& -        
& 0.451        
\\
(c)
& 
&  \multirow{2}{*}{Finetuned}  
& -
& 0.366   
\\
(d)
& 
&  
& \checkmark
& \textbf{0.601}      
\\
\bottomrule
\end{tabular}}
\label{table:ft_strategy}
\end{center}
\end{table}

To mitigate this issue, we implement reset techniques as suggested in \cite{nikishin2022primacy, igl2020nonstataionray}. These techniques involve periodically reinitializing the policy parameters to reintroduce plasticity into the model.
Following \cite{nikishin2022primacy}, we reset the policy parameters for every 40,000 updates.

Table \ref{table:ft_strategy} presents the finetuning performance for different strategies employed in the case of SimTPR. As in \cite{sgi}, simply finetuning the encoder resulted in a relatively low IQM score of 0.366 (Table \ref{table:ft_strategy}.(c)). This score was improved upon when we froze the encoder, leading to an IQM score of 0.451 (Table \ref{table:ft_strategy}.(b)). However, the most significant enhancement was achieved by using the reset strategy with pretrained models. Here, SimTPR achieved an IQM score of 0.601 (Table \ref{table:ft_strategy}.(d)), outperforming the other strategies by a large margin.

In summary, despite the propensity of pretrained representations to overfit to earlier interactions, our results illustrate that pretraining consistently outperforms training from scratch. 
Importantly, the introduction of a periodic reset was able to significantly enhance the fine-tuning performance of these pretrained models.

\section{Conclusion, Limitations, and Future Work}
In this paper, we introduce SimTPR, a novel unsupervised representation learning framework for reinforcement learning. 
By using a feature decorrelation loss to prevent representational collapse, SimTPR efficiently learns temporally predictive representations. Our experimental findings reveal that SimTPR enhances sample efficiency notably on the Atari100k benchmark and attains superior performance in both state and demonstration representation learning.
Furthermore, through a systematic analysis of our framework, we find that the feature rank of the pretrained representations is an important factor affecting the fine-tuning performance.

Despite promising results, our study has a few limitations.
First, our evaluations are confined to Atari environments.
Future research could explore its applicability in diverse environments such as Deepmind Control Suite \cite{tassa2018dmc} or Procedural Generation \cite{cobbe2020procgen}.

Second, we explored a simple variant for fine-tuning, the reset mechanism  \cite{nikishin2022primacy}.
Despite its simplicity, the reset significantly improves the performance of a naive fine-tuning strategy.
Future work should broaden this exploration to include other fine-tuning techniques (e.g., self-supervised objectives, knowledge distillation), to effectively leverage the pretrained representations at fine-tuning.

Last, our observation on the importance of feature rank lacks theoretical validation.
Recent research emphasizes the need for maintaining model plasticity for sample-efficient RL, which refers to its adaptability to new targets \cite{lyle2023understanding_plasticity}.
They discovered that the factors such as the number of active units \cite{abbas2023loss}, the smoothness of the loss landscape \cite{lyle2023understanding_plasticity}, and the feature rank \cite{lyle2022capacity} are potential contributors to model plasticity. 
Further investigating these factors can guide the design of better pretraining objectives that generate generalizable and plastic representations.

\section{Acknowledgement}

This work was supported by Institute of Information \& communications Technology Planning \& Evaluation (IITP) grant funded by the Korea government(MSIT) (No.2021-0-02068, Artificial Intelligence Innovation Hub), No. 2020-0-00368, A Neural-Symbolic Model for Knowledge Acquisition and Inference Techniques), and the National Research Foundation of Korea (NRF) grant funded by the Korea government (MSIT) (No. NRF-2022R1A2B5B02001913).
We also thank Kakao Enterprise for its computational support with GPU Cloud.



\bibliography{main}
\bibliographystyle{icml2023}

\newpage
\appendix
\onecolumn

\section{Implementation Details}
\label{appendix:impl_details}

\subsection{Pretraining}
\label{appendix:impl_pretrain}

In this section, we describe the pretraining details of SimTPR.
First, we explain the architectural details of four major components of our model:  convolutional encoder, projector, transition model, and predictor.

\textbf{Convolutional encoder: $f$.} 
A convolutional encoder is used to encode the augmented states into latent representations. 
The encoder is composed of a 30-layer convolutional network, based on the design from \cite{sgi}.
The encoder is largely divided into three blocks, where each block comprises of one downsample residual block and two residual blocks.
Each residual block follows the design from EfficientNet \cite{tan2019efficientnet}, which is three convolutional layers with an inverted bottleneck structure.
The three layers are composed of a convolutional layer with 1x1 filters, a group convolutional layer with 3x3 filters that increases the number of channels by double the input channel, and a convolutional layer with 1x1 filters with output channels.
Batch normalization and ReLU activation are applied between these convolutional layers.
The spatial resolution of the encoder's output is $(7, 7, 3136)$.

\textbf{Projector: $g$.} Projector maps the encoded representations to the $d$-dimensional latent space ($d=512$).
We compose the projector as an MLP layer with 512 hidden dimensions where ReLU activation is in-between.

\textbf{Transition model: $h$.} Transition model maps the output of the projector into latent representations of future states. 
We utilized a casual transformer, based on the design from ~\cite{radford2018gpt}.
The casual transformer is composed of 2 casual transformer blocks, each with 512 input dimensions, 8 heads, and 2048 hidden dimensions.

\textbf{Predictor: $q$.} Predictor maps the output of the transition model to the $d$-dimensional latent space ($d=512$). 
The predictor is composed of an MLP layer with 512 hidden dimensions where ReLU activation and batch normalization are in-between. 

In pretraining, we use the same hyperparameter of the state and demonstration pretraining where the only difference is the use of action prediction weight in the demonstration dataset. For simplicity, we use the action prediction weight as $\lambda_a = 1.0$.
Detailed hyperparameters are described in Table \ref{table:pt_hyperparameter}. On a single A100 GPU, the pretraining takes 1 to 2 days for each run. 

\begin{table}[h]
\vspace{-3mm}
\begin{center}
\caption{Hyperparameters of SimTPR for pretraining on DQN replay dataset.}
\vspace{2mm}
\resizebox{0.65 \textwidth}{!}{
\begin{tabular}{lr}
\toprule
Hyperparameter & Value \\
\midrule \\[-2.5ex]
State size               & (84, 84, 4) \\
Grey scaling             & True \\
Data Augmentation        & Random Shifts and Intensity Jittering  \\
Random Shifts            & $\pm$ 4 pixels \\ 
Intensity Jiterring scale& 0.05 \\
\midrule \\[-2.5ex]
Channels for each block                 & (16, 64, 64)  \\
Strides for each block                  & (3, 2, 2) \\
Number of future steps to associate (k) & 1     \\
Off-diagonal weight $(\lambda_o)$       & 0.005 \\
Decorrelation weight $(\lambda_d)$      &  0.01   \\
Action prediction weight $(\lambda_a)$  &   1      \\
\midrule \\[-2.5ex]
Epochs                   & 100 \\
Batch size               & 640  \\
Number of Sequences (N)  & 64  \\
Sequence length (T)      & 10  \\
Optimizer ($\beta_1$, $\beta_2$, $\epsilon$) 
                         & AdamW (0.9, 0.999, 0.000015) \\
Learning rate            & 0.0003  \\
Weight decay             & 0.00001  \\
Max gradient norm        & 0.5 \\
\bottomrule
\end{tabular}}
\label{table:pt_hyperparameter}
\end{center}
\vspace{-3mm}
\end{table}

\newpage

\subsection{Linear Probing}
\label{appendix:linear_probe}

Following the evaluation protocol from \cite{lightweight_probe}, a linear classifier was trained on top of the frozen encoder and predicted the reward and action for each state.
To evaluate the effectiveness of the pretrained representations in capturing the policy and value function of experts, we utilized the last 100,000 frames from the 50 million frame DQN replay dataset  \cite{agarwal2020optimistic} with a 4:1 train/eval split.

To simplify the reward probing task, we formulated it as a binary classification problem where the objective is to predict the occurrence of a reward for a given state.
We trained a logistic regression model using the majorization-minimization optimization (MISO) algorithm \cite{mairal2015miso} for a total of 300 iterations.

For the action probing task, we train a multi-class classifier with a softmax focal loss \cite{lin2017focal} to address the class imbalance of the action labels. 
The model was trained for 50 epochs using the SGD optimizer with a learning rate of $0.2$, batch size of $256$, and a weight decay of $10^{-6}$.
For each epoch, the learning rate was decayed by a step-wise scheduler with a step size of $10$ and gamma of $0.1$.

Both tasks were performed on an A100 GPU and completed within 5 minutes.

\subsection{Finetuning}

Following the evaluation protocol from \cite{lightweight_probe, sgi}, we focus on the Atari 100k benchmark \cite{kaiser2019model}, where only 100k steps of interactions are allowed at the finetuning phase.
We train a Q-learning head on top of the frozen encoder using the Rainbow algorithm \cite{hessel2018rainbow}.
Here we do not apply any auxiliary self-supervised loss and rely on noisy layers for exploration.
A detailed hyperparameter for the finetuning setup is described in Table \ref{table:ft_hyperparameter}.
On an A100 GPU, each run takes from 2 to 3 hours.

\begin{table}[h]
\vspace{-3mm}
\begin{center}
\caption{Hyperparameters for finetuning on the Atari100k benchmark.}
\vspace{2mm}
\resizebox{0.55 \textwidth}{!}{
\begin{tabular}{lr}
\toprule
Hyperparameter & Value \\
\midrule \\[-2.5ex]
State downsample size    & (84, 84) \\
Grey scaling             & True \\
Data augmentation        & Random Shifts and Intensity Jittering  \\
Random Shifts            & $\pm$ 4 pixels \\ 
Intensity Jiterring scale& 0.05 \\
Frame skip               & 4 \\
Stacked frames           & 4 \\
Action repeat            & 4 \\
\midrule \\[-2.5ex]
Training steps           & 100k \\
Update                   & Distributional Q \\
Dueling                  & True \\
Support of Q-distribution& 51 \\
Discount factor $\gamma$ & 0.99 \\
Batch size               & 32  \\
Optimizer ($\beta_1$, $\beta_2$, $\epsilon$) 
                         & Adam (0.9, 0.999, 0.000015) \\
Learning rate            & 0.00003 \\
Max gradient norm        & 10 \\
Priority exponent        & 0.5 \\
Priority correction      & 0.4 $\rightarrow$ 1 \\
Exploration              & Noisy nets \\
Noisy nets parameter     & 0.5 \\
Replay buffer size       & 100k \\
Min buffer size for sampling & 2000 \\
Replay per training step & 1 \\
Updates per replay step  & 2 \\
Multi-step return length & 10 \\
Q-head hidden units      & 512 \\
Q-head non-linearity     & ReLU \\
Evaluation trajectories  & 50 \\
\bottomrule
\end{tabular}}
\label{table:ft_hyperparameter}
\end{center}
\vspace{-3mm}
\end{table}

\newpage

\section{Baseline Implementations}
\label{appendix:baselines}

Our comparison in Table \ref{table: main_result} is based on our reproduction of the related methods with the exception of SGI and BarlowBalance.
We have implemented the competitors as faithfully as possible following each individual paper.
In order to ensure fairness in comparison, we have employed the same convolutional encoder, latent dimension, data augmentation, pretraining data, number of pretraining epochs, and the optimizer (AdamW) across all methods.
For each method, we have varied the learning rate from $\{0.001, 0.0003, 0.0001\}$ during pretraining and selected the model with the highest linear probing score for each method.
Then, we finetuned the selected models for 10 random seeds and report the finetuning performance. 
A detailed description of the training setup for each method is provided in the following section.

\subsection{VAE}
For the variational autoencoder (VAE), data augmentation was applied to each state, followed by the convolutional encoder to obtain the latent representations.
 Subsequently, the mean ($\mu$) and log variance ($\log \sigma^2$) of the variational posterior of VAE were obtained by the MLP layer with 256 hidden dimensions, respectively.
Then, the random variable $z$ was sampled from the variational posterior $N(z;\mu, \sigma^2I)$  and used to reconstruct the original states through the convolutional decoder. 
The architecture of the convolutional decoder is symmetric to the convolutional encoder as \cite{long2015fcn}.
We trained VAE by jointly minimizing the KL divergence between the variational posterior and the prior $N(0, I)$, and the Mean Squared Error (MSE) between the reconstructed states and the original states.
To balance the loss between two terms, we weighted the KL divergence loss by a factor of $\lambda_{KL} = 0.02$.
To ensure compatibility with a batch size of 640, we set a number of sequences of $N = 640$, with a sequence length of $T = 1$.

\subsection{BarlowTwins}
BarlowTwins learns the representations by decorrelating the features of the augmented states. 
BarlowTwins uses siamese architecture where two differently augmented states are processed by an encoder followed by a projector.
Then, the projected representations, $Z$, and $Z'$ are obtained through this siamese architecture. 
Note that the convolutional encoder and projector are identical to our configuration.
The model is trained to standardize the cross-correlation matrix between projected representations $Z$ and $Z'$ to the identity matrix. 
To balance the on-diagonal and off-diagonal terms, $\lambda_{Barlow}=0.05$ is applied to weight the off-diagonal terms of the cross-correlation matrix.
To ensure compatibility with a batch size of 640, we utilize a number of sequences, $N = 640$, with a sequence length of $T = 1$.

\subsection{CURL}
Contrastive Unsupervised Representations for Reinforcement Learning (CURL) employs a siamese model, where the convolutional encoder and a projection layer are identical to our setup. 
The siamese model is coupled with the Momentum Encoder, with a momentum coefficient of $\tau = 0.99$.
Two different data augmentations are applied to each state, followed by the siamese model to obtain $Z$ and $Z'$. 
A prediction layer, consisting of a residual MLP layer with 512 hidden dimensions, is applied to $Z$ to obtain the prediction $Q$.
Then, CURL is trained to minimize the contrastive loss between $Q$ and $Z'$, with a temperature of 0.1. 
For a fair comparison, the contrastive loss is also computed symmetrically between $Q'$ and $Z$.
To ensure compatibility with a batch size of 640, we utilized a number of sequences as $N = 640$, with a sequence length of $T = 1$.

\subsection{RSSM}
Recurrent State-Space Model (RSSM) learns temporally predictive representations by predicting future states in a latent space and reconstructing them in the original state space. 
To begin with, data augmentation was applied to each state, followed by the convolutional encoder to obtain the latent representations.
Let $s_t$ be the latent representation of the state at time step $t$,
We explain the architectural details of four major components of RSSM: the recurrent model, the transition model, the representation model, and the image decoder, as follows:
\begin{equation}\label{eqn: RSSM}
\begin{aligned}
    \text{Recurrent model:} \: h_{t+1} &= f_\theta(h_t, z_t),
     \\
    \text{Transition model:} \: \hat{z}_{t+1} &= p_\phi(\hat{z}_{t+1}\;|\;h_{t+1}),
     \\
     \text{Representation model:} \: z_{t+1} &= q_\phi(z_{t+1}\;|\;h_{t+1}, s_{t+1}),
     \\
     \text{Image Decoder:} \: \hat{s}_{t+1} &= f_\phi(\hat{s}_{t+1}\;|\;h_{t+1}, z_{t+1}).
\end{aligned}
\end{equation}

\newpage

\textbf{Recurrent model.} Recurrent model predicts the latent of the next states and consists of an MLP layer with 512 hidden dimensions, ReLU activation, and a gated recurrent unit (GRU) cell with 512 hidden dimensions. 

\textbf{Transition model.} Transition model serves as a prior and consists of a Linear layer with hidden dimensions of 512, ReLU activation, an output Linear layer with 256 output dimensions, and a Softplus activation to obtain mean $\mu$ and standard deviation $\sigma$ with 128 hidden dimensions, respectively.

\textbf{Representation model.} Representation model serves as a posterior and is identical to the structure of the transition model.

\textbf{Image decoder.} Image decoder reconstructs the target state through the convolutional decoder.
The architecture of the convolutional decoder is symmetric to the convolutional encoder (Long et al., 2015), which is identical to the architecture used in VAE.

RSSM was trained to minimize the KL divergence between the representation model and the transition model, as well as the Mean Squared Error (MSE) between the reconstructed states and the target states. 
In order to balance the loss between the KL divergence and MSE error, we weigh the KL divergence loss by  a factor of $\lambda_{KL} = 0.025$.
To ensure compatibility with a batch size of 640, we used a total of $N=64$ sequences, each with a sequence length of $T=10$.

\subsection{ATC}
Augmented Temporal Contrast (ATC) learns temporally predictive representations by maximizing the similarity between the representations of current states and states with a short time difference.
ATC employs a siamese model,  where the convolutional encoder and a projection layer are identical to our setup. 
The siamese model is coupled with the Momentum Encoder, utilizing a momentum coefficient of $\tau = 0.99$.
In training, two different data augmentations are applied to each state, followed by the siamese model to obtain $Z$ and $Z'$. 
A prediction layer, which is also identical to our configuration is applied to $Z$ to obtain the prediction $Q$.
Then, a contrastive loss is applied, which maximizes the similarity between predictions from future states with $k$-steps distances (i.e., the similarity between $Q_t$ and $Z'_{t+k}$ at time step $t$).
Following \cite{atc, sgi}, we set $k=3$.
For a fair comparison, we symmetrize the loss function and set a number of sequences of $N=64$, with a sequence length of $T=10$.

\subsection{BC} 

For behavioral cloning, we applied data augmentation to each state and trained a prediction model to predict the action for each state.
The action prediction layer is applied to the output of the convolutional encoder which is implemented as a two-layer MLP with $512$ hidden dimensions, where a ReLU activation is applied in between.
To ensure compatibility with the batch size of $640$, we utilized a number of sequences of $N=640$ and a sequence length of $T=1$.
Through empirical analysis, we found that behavioral cloning serves as a strong baseline among the demonstration pretraining methods.

\subsection{IDM}

Inverse Dynamics Modeling (IDM) is trained to predict the action between two consecutive states.
Similar to behavioral cloning, the action prediction layer is implemented as a two-layer multi-layer perceptron (MLP) with $512$ hidden dimensions and ReLU activation is applied in between.
In the training process, we first applied data augmentation to two consecutive states and used a convolutional encoder to obtain the latent representations for each state.
The resulting representations were concatenated and processed by the action predictor to predict the action between the states.
In order to ensure compatibility with the batch size of $640$, we utilized a number of sequences of $N=320$ and a sequence length of $T=2$.
\newpage

\section{PseudoCode}

\subsection{State Pretraining}
\begin{algorithm}[h]
    \caption{PyTorch-like Pseudocode of SimTPR for State Pretraining}
    \PyComment{f: encoder} \\
    \PyComment{g: projector} \\
    \PyComment{h: transition model} \\
    \PyComment{p: predictor} \\
    \\
    \PyComment{n: batch size} \\
    \PyComment{t: sequence length} \\
    \PyComment{d: dimension of the latent representations} \\
    \PyComment{k: future steps to associate with} \\
    \\
    \PyDef{for} \PyCode{s} \PyDef{in} \PyCode{loader:}    \PyComment{load a batch of (n,t) samples}
    
    \Indp 
    \PyComment{compute embeddings} \\
    \PyCode{x1, x2 = aug(s), aug(s)} \\
    \PyCode{z1, z2 = g(f(x1)), g(f(x2))} \PyComment{projections, (n,t,d)} \\
    \PyCode{q1, q2 = p(h(z1), p(h(z2))}  \PyComment{predictions, (n,t,d)} \\

    \PyComment{l2-normalize} \\
    \PyCode{z1, z2 = normalize(z1, dim=2), normalize(z2, dim=2)} \\
    \PyCode{q1, q2 = normalize(q1, dim=2), normalize(q2, dim=2)} \\

    \PyComment{compute loss} \\
    \PyCode{L\_sim = distance(q1, z2.detach())/2 + distance(q2, z1.detach())/2)} \\
    \PyCode{L\_decorr = standardize(z1, z2) } \\
    \PyCode{L = L\_sim + lambda\_d.mul(L\_decorr)} \\

    \PyComment{optimize} \\
    \PyCode{L.backward() } \\
    \PyCode{optimizer.step()} \\

    \Indm

\PyDef{def} \PyCode{distance(q, z):}

    \Indp   

    \PyCode{z1, z2 = z1[:, k:], z2[:, k:]}   \PyComment{(n,t-k,d)} \\
    \PyCode{q1, q2 = q1[:, :-k], q2[:, :-k]} \PyComment{(n,t-k,d)} \\

    \PyDef{return} \PyCode{(q1 - z2).pow(2).sum(dim=2).mean()} \\
    
    \Indm

\PyDef{def} \PyCode{standardize(z1, z2):}

    \Indp   

    \PyComment{reshape} \\
    \PyCode{z1, z2 = z1.reshape(n*t,d), z2.reshape(n*t,d)} \\
    \PyCode{q1, q2 = q1.reshape(n*t,d), q2.reshape(n*t,d)} \\

    \PyComment{compute cross-correlation matrix (d,d)} \\
    \PyCode{z1\_norm = (z1 - z1.mean(0)) / z\_1.std(0)} \\
    \PyCode{z2\_norm = (z2 - z2.mean(0)) / z\_2.std(0)} \\
    \PyCode{c = mm(z1\_norm.T, z2\_norm)/n} \\

    \PyComment{loss} \\
    \PyCode{c\_on =on\_diagonal(c)} \PyComment{(d)} \\
    \PyCode{c\_off =off\_diagonal(c)} \PyComment{(d(d-1))}\\
    \PyDef{return} \PyCode{(c\_on - ones(d)).sum() + lambda\_o.mul(c\_off.sum())}

    \Indm
    
\label{algo:SimTPR_video}
\end{algorithm}

\newpage

\subsection{Demonstration Pretraining}

\begin{algorithm}[h]
    \caption{PyTorch-like Pseudocode of SimTPR for Demonstration Pretraining}
    \PyComment{f: encoder} \\
    \PyComment{g: projector} \\
    \PyComment{h: transition model} \\
    \PyComment{p: predictor} \\
    \PyComment{r: action predictor} \\
    \\
    \PyComment{n: batch size} \\
    \PyComment{t: sequence length} \\
    \PyComment{d: dimension of the latent representations} \\
    \PyComment{k: future steps to associate with} \\
    \PyComment{n\_a: a number of action in the environment} \\
    \\
    \PyComment{distance(q, z): a distance function defined in algorithm \ref{algo:SimTPR_video}} \\
    \PyComment{standardize(z1, z2): a standardization function defined in algorithm \ref{algo:SimTPR_video}} \\
    \PyComment{cross\_entropy(y, a): a standard cross-entropy loss where y is the un-normalized logits} \\
    \\
    \PyDef{for} \PyCode{s, a} \PyDef{in} \PyCode{loader:}    \PyComment{load a batch of (n,t) samples}
    
    \Indp 
     \PyComment{compute embeddings} \\
    \PyCode{x1, x2 = aug(s), aug(s)} \\
    \PyCode{z1, z2 = g(f(x1)), g(f(x2))} \PyComment{projections, (n,t,d)} \\
    \PyCode{y = embed(a)} \PyComment{linearly embed actions, (n,t,d)} \\
    \PyCode{tau1 = alternate\_concat(z1, y)} \PyComment{tau1 = (z1\_1,a\_1,...,z1\_t,a\_t), (n,2*t,d)}\\
    \PyCode{tau2 = alternate\_concat(z2, y)} \PyComment{tau2 = (z2\_1,a\_1,...,z2\_t,a\_t), (n,2*t,d)}\\
    \PyCode{c1, c2 = h(tau1), h(tau2))}  \PyComment{transition, (n,2*t,d)} \\
    \PyCode{q1, q2 = p(c1[:,1,3,...,2t-1]), p(c2[:,1,3,...,2t-1]))}  \PyComment{latent prediction, (n,t,d)} \\
    \PyCode{l1, l2 = r(c1[:,0,2,...,2t-2]), r(c2[:,0,2,...,2t-2]))}  \PyComment{action prediction, (n,t,n\_a)} \\

    \PyComment{l2-normalize} \\
    \PyCode{z1, z2 = normalize(z1, dim=2), normalize(z2, dim=2)} \\
    \PyCode{q1, q2 = normalize(q1, dim=2), normalize(q2, dim=2)} \\

    \PyComment{compute loss} \\
    \PyCode{L\_sim = distance(q1, z2.detach())/2 + distance(q2, z1.detach())/2)} \\
    \PyCode{L\_decorr = standardize(z1, z2) } \\
    \PyCode{L\_act = cross\_entropy(l1, a)/2 + cross\_entropy(l2, a)/2 }\\
    \PyCode{L = L\_sim + lambda\_d.mul(L\_decorr) + lambda\_a.mul(L\_act)} \\

    \PyComment{optimize} \\
    \PyCode{L.backward() } \\
    \PyCode{optimizer.step()} \\
    
    \Indm
    
\label{algo:SimTPR_demon}
\end{algorithm}

\newpage

\section{Uncertainty-Aware Comparison}

\subsection{State Representation Learning}

\begin{figure}[h]
\begin{center}
\includegraphics[width=0.9\linewidth]{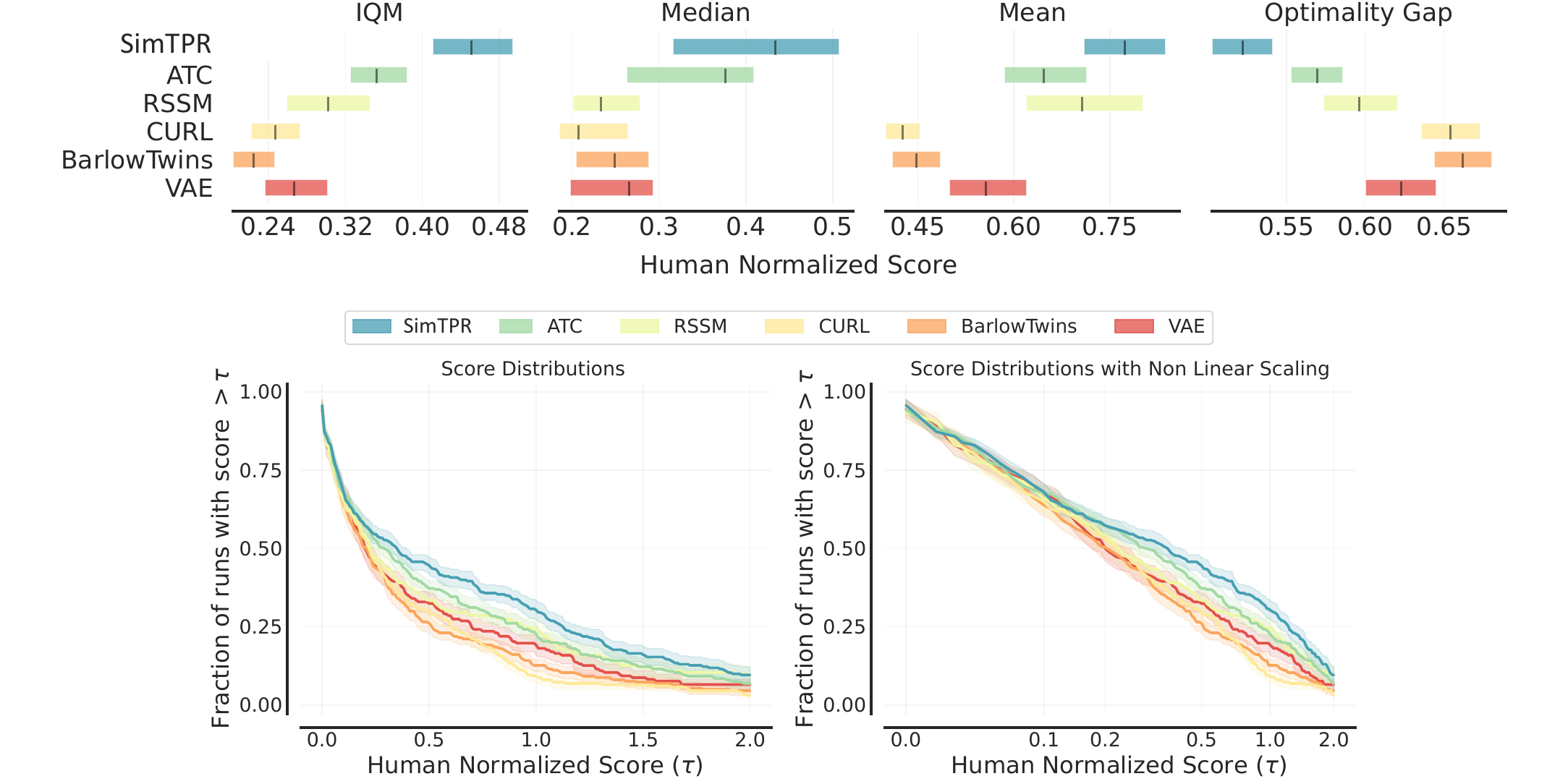}
\end{center}
\vspace{-3mm}
\caption{\textbf{Uncertainty-aware comparison for State Representation Learning.} \textbf{Top.} 95\% Confidence Intervals for IQM, Median, Mean and Optimality Gap scores for each state representation learning method in Table \ref{table: main_result}. 
\textbf{Bottom.} Average score distributions for each method where shaded regions represent a pointwise 95\% confidence interval.
The confidence intervals are estimated using the percentile bootstrap with stratified sampling as \cite{agarwal2021iqm}.}
\label{fig:state_uncertainty}
\end{figure}

\subsection{Demonstration Representation Learning}

\begin{figure}[h]
\begin{center}
\includegraphics[width=0.9\linewidth]{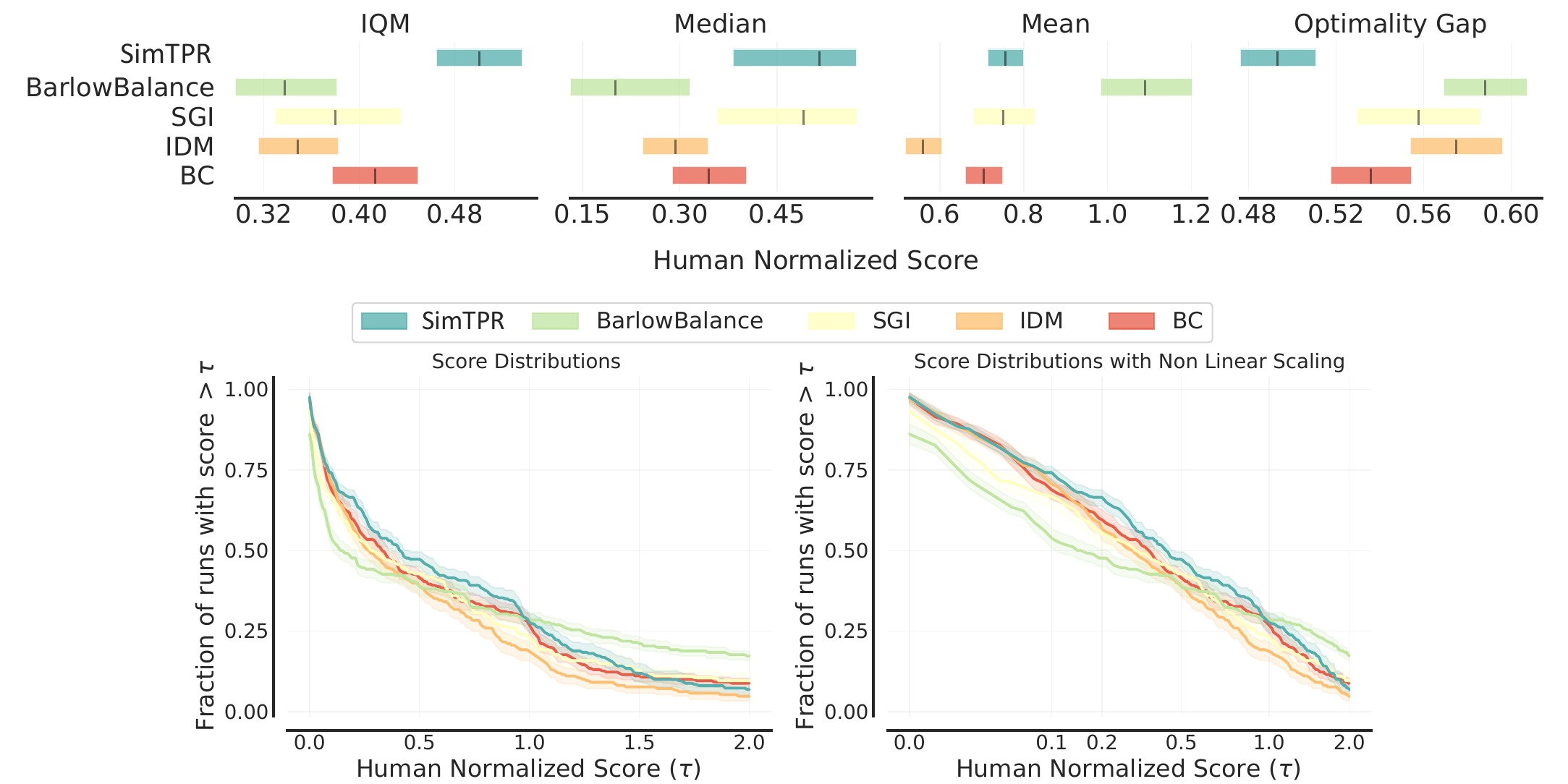}
\end{center}
\vspace{-3mm}
\caption{\textbf{Uncertainty-aware comparison for Demonstration Representation Learning.} \textbf{Top.} 95\% Confidence Intervals for IQM, Median, Mean and Optimality Gap scores for each demonstration representation learning method in Table \ref{table: main_result}. 
\textbf{Bottom.} Average score distributions for each method where shaded regions represent a pointwise 95\% confidence interval.
The confidence intervals are estimated using the percentile bootstrap with stratified sampling as \cite{agarwal2021iqm}.}
\label{fig:demon_uncertainty}
\vspace{-3cm}
\end{figure}

\newpage

\section{Does Higher Linear Probing Score indicates a better finetuning Performance?}

\begin{figure}[h]
\begin{center}
\includegraphics[width=0.7\linewidth]{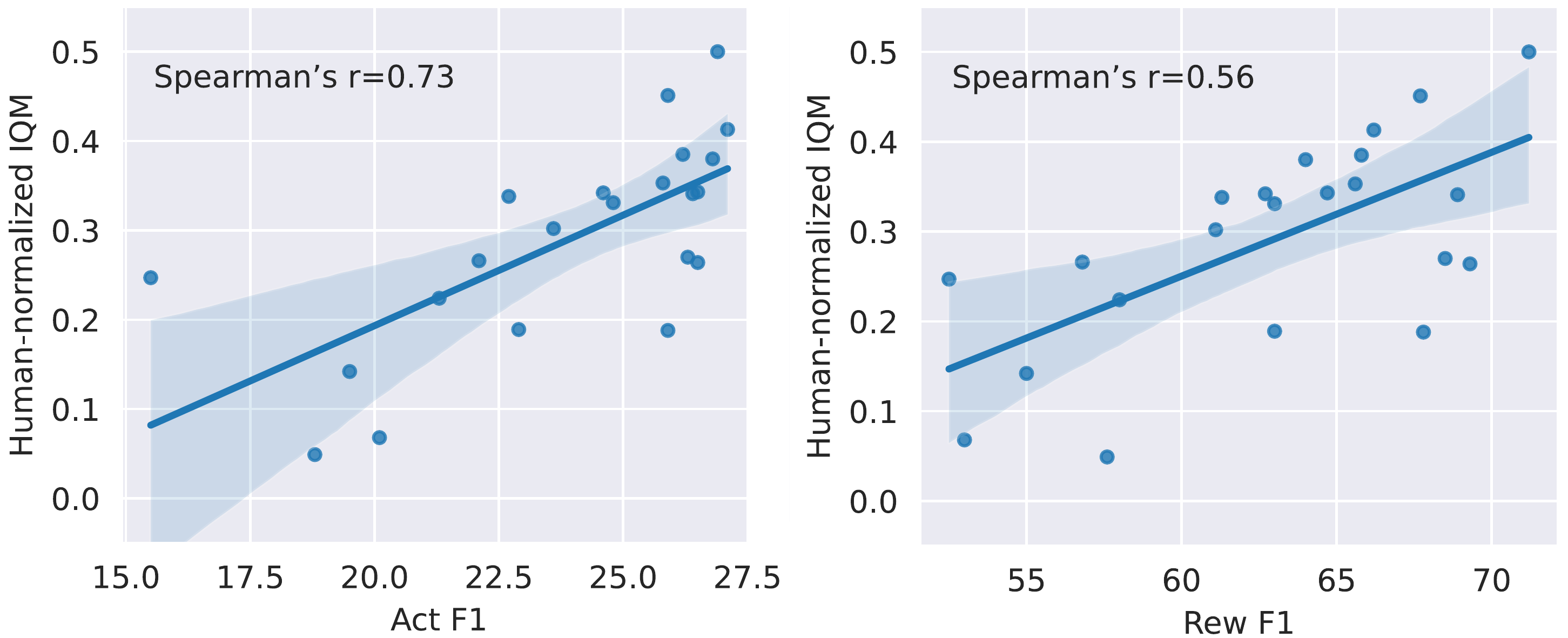}
\end{center}
\vspace{-3mm}
\caption{\textbf{Correlation between the Linear Probing Score and the finetuning Performance.} \textbf{Left.} Correlation between the Act F1 score and IQM of the HNS. \textbf{Right.} Correlation between the Rew F1 score and IQM of the HNS. } 
\label{fig:state_uncertainty}
\vspace{-3mm}
\end{figure}

In order to assess the representation quality of a pretrained model, a commonly employed protocol is to train reinforcement learning (RL) algorithms on top of the pretrained model. 
However, this approach can be computationally intensive and may result in high-variance outcomes. 
To address this problem, a recent study by \cite{lightweight_probe} has proposed the use of two linear probing tasks as an alternative method for evaluating representation quality. 
These tasks, action prediction, and reward prediction are designed to predict the actions and rewards of experts and are computationally efficient with low variance.
Empirically, a study by \cite{lightweight_probe} has shown that these probing scores have a high correlation to the RL agent's finetuning score.

In this study, we aim to investigate the relationship between linear probing scores and finetuning performance for reinforcement learning (RL) agents.
We present an analysis of the correlation between the linear probing scores (Act F1 and Rew F1) and the finetuning performance (IQM of human-normalized score) for the RL agent in Figure \ref{fig:state_uncertainty}.
Each data point in the figure represents a different pretraining method, including ablation studies of SimTPR and various baselines.
To evaluate the correlation, we applied Spearman's rank correlation test to the data.
The results of this analysis indicate a moderate correlation between Action F1 and IQM, with a Spearman's correlation coefficient of 0.73, and a correlation coefficient between the Rew F1 and IQM of 0.56.
Thus, based on the moderate correlation, it may not be sufficient to conclude that a higher linear probing score guarantees a better finetuning performance for the RL agent.

\clearpage

\section{Full Results on Atari100k}

\subsection{State Representation Learning}

\begin{table}[h]
\begin{center}
\caption{\textbf{Mean episodic scores of state representation learning.} We report the mean episodic scores on the 26 Atari games for state representation learning methods. The results are recorded at the end of training and averaged over 10 random seeds.
SimTPR outperforms the other methods on all aggregate metrics.}
\vspace{0.8ex}
\resizebox{0.98\textwidth}{!}{
\begin{tabular}{l  c cccccccc}
\toprule &\\[-2.5ex]
Game       & Random & Human & DrQ & VAE & BarlowTwins & CURL & RSSM & ATC & SimTPR \\[0.2ex]
\hline \\[-2ex]
Alien          & $227.8$    & $7127.7$   & $798.52$   & $1013.74$  & $853.5$    & $823.56$   & $1013.23$  & $845.84$   & $784.12$   \\
Amidar         & $5.8$      & $1719.5$   & $123.91$   & $250.49$   & $157.44$   & $128.49$   & $191.57$   & $124.87$   & $158.35$   \\
Assault        & $222.4$    & $742.0$    & $534.88$   & $783.52$   & $635.76$   & $616.87$   & $860.19$   & $716.34$   & $853.58$   \\
Asterix        & $210.0$    & $8503.3$   & $605.5$    & $560.86$   & $528.6$    & $538.6$    & $537.5$    & $574.7$    & $521.4$    \\
BankHeist      & $14.2$     & $753.1$    & $124.72$   & $56.57$    & $57.18$    & $91.96$    & $35.37$    & $395.7$    & $402.14$   \\
BattleZone     & $2360.0$   & $37187.5$  & $10714.0$  & $3274.29$  & $5032.0$   & $6718.0$   & $6746.67$  & $9168.0$   & $6796.0$   \\
Boxing         & $0.1$      & $12.1$     & $3.61$     & $32.71$    & $25.23$    & $10.01$    & $41.66$    & $32.76$    & $42.13$    \\
Breakout       & $1.7$      & $30.5$     & $11.17$    & $4.07$     & $14.53$    & $10.83$    & $3.29$     & $21.45$    & $55.67$    \\
ChopperCommand & $811.0$    & $7387.8$   & $914.4$    & $736.57$   & $873.8$    & $1012.8$   & $829.67$   & $836.4$    & $851.6$    \\
CrazyClimber   & $10780.5$  & $35829.4$  & $11203.6$  & $32358.86$ & $20787.6$  & $26213.0$  & $49406.67$ & $69789.0$  & $56273.2$  \\
DemonAttack    & $152.1$    & $1971.0$   & $384.7$    & $471.57$   & $704.84$   & $785.67$   & $801.92$   & $925.24$   & $1484.47$  \\
Freeway        & $0.0$      & $29.6$     & $26.94$    & $15.22$    & $22.29$    & $16.49$    & $25.79$    & $0.0$      & $6.68$     \\
Frostbite      & $65.2$     & $4334.7$   & $607.3$    & $2587.0$   & $1264.02$  & $874.96$   & $2222.07$  & $252.0$    & $268.48$   \\
Gopher         & $257.6$    & $2412.5$   & $344.88$   & $800.06$   & $453.84$   & $628.44$   & $657.67$   & $761.56$   & $1022.44$  \\
Hero           & $1027.0$   & $30826.4$  & $6967.47$  & $7102.73$  & $8206.1$   & $7123.86$  & $5870.07$  & $10773.08$ & $10804.96$ \\
Jamesbond      & $29.0$     & $302.8$    & $225.7$    & $395.0$    & $348.8$    & $312.5$    & $465.17$   & $417.9$    & $298.5$    \\
Kangaroo       & $52.0$     & $3035.0$   & $2907.0$   & $1214.0$   & $819.4$    & $854.6$    & $618.0$    & $1520.0$   & $3227.8$   \\
Krull          & $1598.0$   & $2665.5$   & $3178.16$  & $4532.11$  & $3911.5$   & $4277.88$  & $5332.93$  & $3722.96$  & $4678.62$  \\
KungFuMaster   & $258.5$    & $22736.3$  & $3085.4$   & $17444.29$ & $7982.6$   & $13353.2$  & $10321.67$ & $16078.6$  & $14197.6$  \\
MsPacman       & $307.3$    & $6951.6$   & $843.0$    & $1636.57$  & $1079.96$  & $1180.2$   & $1633.0$   & $1268.54$  & $1249.58$  \\
Pong           & $-20.7$    & $14.6$     & $-14.88$   & $-10.04$   & $-13.33$   & $-4.59$    & $-4.03$    & $0.51$     & $-2.57$    \\
PrivateEye     & $24.9$     & $69571.3$  & $80.0$     & $85.71$    & $90.0$     & $122.33$   & $-43.53$   & $124.84$   & $90.0$     \\
Qbert          & $163.9$    & $13455.0$  & $2337.6$   & $3893.64$  & $2709.65$  & $2971.9$   & $3747.08$  & $3299.7$   & $3091.7$   \\
RoadRunner     & $11.5$     & $7845.0$   & $6439.6$   & $4411.43$  & $8172.2$   & $9915.2$   & $8997.0$   & $9837.2$   & $9780.6$   \\
Seaquest       & $68.4$     & $42054.7$  & $407.16$   & $703.66$   & $352.24$   & $465.88$   & $753.8$    & $517.12$   & $424.96$   \\
UpNDown        & $533.4$    & $11693.2$  & $3198.66$  & $11897.83$ & $4900.44$  & $2404.42$  & $20043.8$  & $15108.64$ & $16623.82$ \\
\midrule \\[-2.7ex]
IQM        & $0.0$  & $1.0$  & $0.161$ & $0.266$ & $0.224$ & $0.247$ & $0.302$ & $0.353$ & $\textbf{0.451}$ \\
Median     & $0.0$  & $1.0$  & $0.157$ & $0.266$ & $0.249$ & $0.208$ & $0.235$ & $0.376$ & $\textbf{0.434}$ \\
Mean       & $0.0$  & $1.0$  & $0.308$ & $0.556$ & $0.448$ & $0.426$ & $0.707$ & $0.647$ & $\textbf{0.773}$ \\
OG         & $0.0$  & $1.0$  & $0.724$ & $0.623$ & $0.662$ & $0.654$ & $0.595$ & $0.570$ & $\textbf{0.522}$ \\
\bottomrule 
\end{tabular}}
\end{center}
\label{table:state_full_result}
\end{table}

\newpage

\subsection{Demonstration Representation Learning}

\begin{table}[h]
\begin{center}
\caption{\textbf{Mean episodic scores of demonstration representation learning.} We report the mean episodic scores on the 26 Atari games for demonstration representation learning methods. The results are recorded at the end of training and averaged over 10 random seeds.
SimTPR outperforms the other methods on all aggregate metrics except Mean.}
\vspace{0.8ex}
\resizebox{0.8\textwidth}{!}{
\begin{tabular}{l  c cccccc}
\toprule &\\[-2.5ex]
Game       & Random & Human &BC & IDM & SGI & BarlowBalance & SimTPR \\[0.2ex]
\hline \\[-2ex]
Alien          & $227.8$    & $7127.7$   & $837.24$   & $1007.2$   & $1091.38$  & $981.44$   & $1006.1$   \\
Amidar         & $5.8$      & $1719.5$   & $143.47$   & $134.97$   & $233.47$   & $170.37$   & $127.74$   \\
Assault        & $222.4$    & $742.0$    & $735.42$   & $535.04$   & $569.69$   & $903.22$   & $1053.48$  \\
Asterix        & $210.0$    & $8503.3$   & $709.0$    & $587.57$   & $398.75$   & $436.0$    & $643.8$    \\
BankHeist      & $14.2$     & $753.1$    & $363.18$   & $557.83$   & $397.33$   & $205.38$   & $449.04$   \\
BattleZone     & $2360.0$   & $37187.5$  & $7838.0$   & $7937.14$  & $2920.0$   & $3812.0$   & $9014.0$   \\
Boxing         & $0.1$      & $12.1$     & $41.93$    & $22.52$    & $44.41$    & $62.13$    & $13.01$    \\
Breakout       & $1.7$      & $30.5$     & $34.61$    & $24.47$    & $13.74$    & $1.11$     & $40.56$    \\
ChopperCommand & $811.0$    & $7387.8$   & $1124.6$   & $931.14$   & $815.0$    & $658.4$    & $829.6$    \\
CrazyClimber   & $10780.5$  & $35829.4$  & $62124.0$  & $45292.86$ & $43783.5$  & $94975.4$  & $47558.8$  \\
DemonAttack    & $152.1$    & $1971.0$   & $1587.94$  & $748.46$   & $182.7$    & $343.5$    & $2571.99$  \\
Freeway        & $0.0$      & $29.6$     & $21.4$     & $16.75$    & $18.53$    & $19.27$    & $25.65$    \\
Frostbite      & $65.2$     & $4334.7$   & $385.94$   & $540.91$   & $1207.2$   & $638.76$   & $245.38$   \\
Gopher         & $257.6$    & $2412.5$   & $994.8$    & $751.89$   & $1253.85$  & $1149.36$  & $1241.48$  \\
Hero           & $1027.0$   & $30826.4$  & $8232.99$  & $7398.11$  & $4336.45$  & $5302.16$  & $9986.87$  \\
Jamesbond      & $29.0$     & $302.8$    & $241.4$    & $318.29$   & $307.38$   & $414.0$    & $419.5$    \\
Kangaroo       & $52.0$     & $3035.0$   & $832.2$    & $630.86$   & $2195.0$   & $2213.0$   & $664.8$    \\
Krull          & $1598.0$   & $2665.5$   & $4809.3$   & $4825.6$   & $6313.55$  & $5551.18$  & $6307.36$  \\
KungFuMaster   & $258.5$    & $22736.3$  & $12934.0$  & $12198.57$ & $14343.25$ & $14359.6$  & $13142.0$  \\
MsPacman       & $307.3$    & $6951.6$   & $1313.14$  & $1356.29$  & $1600.2$   & $415.06$   & $1849.7$   \\
Pong           & $-20.7$    & $14.6$     & $11.35$    & $7.41$     & $3.26$     & $3.51$     & $9.29$     \\
PrivateEye     & $24.9$     & $69571.3$  & $90.0$     & $71.67$    & $61.5$     & $60.2$     & $100.2$    \\
Qbert          & $163.9$    & $13455.0$  & $2754.0$   & $2998.21$  & $714.06$   & $185.5$    & $4576.75$  \\
RoadRunner     & $11.5$     & $7845.0$   & $10749.4$  & $7755.14$  & $7995.25$  & $21362.6$  & $10694.2$  \\
Seaquest       & $68.4$     & $42054.7$  & $487.88$   & $559.77$   & $561.75$   & $579.92$   & $584.2$    \\
UpNDown        & $533.4$    & $11693.2$  & $4414.98$  & $5744.23$  & $18800.9$  & $74548.54$ & $9086.26$  \\
\midrule \\[-2.7ex]
IQM        & $0.0$  & $1.0$  & $0.413$ & $0.343$ & $0.380$ & $0.338$ & $\textbf{0.500}$ \\
Median     & $0.0$  & $1.0$  & $0.345$ & $0.279$ & $0.490$ & $0.201$ & $\textbf{0.515}$ \\
Mean       & $0.0$  & $1.0$  & $0.705$ & $0.564$ & $0.751$ & $\textbf{1.089}$ & $0.757$ \\
OG         & $0.0$  & $1.0$  & $0.536$ & $0.579$ & $0.558$ & $0.588$ & $\textbf{0.493}$ \\
\bottomrule 
\end{tabular}}
\end{center}
\label{table:state_full_result}
\end{table}



\end{document}